%% file: emnlp2022.tex
\pdfoutput=1
\documentclass[11pt]{article}

\usepackage[]{EMNLP2022}

\usepackage{times}
\usepackage{latexsym}

\usepackage[T1]{fontenc}
\usepackage[utf8]{inputenc}
\usepackage{graphicx}  
\usepackage{microtype}

\usepackage{inconsolata}
\usepackage{times}
\usepackage{latexsym}
\usepackage{hyperref}
\usepackage{subcaption}
\usepackage{xcolor}
\usepackage{multirow}

\newcommand{\datasetname}{\textsc{NewsClaims}}
\newcommand{\subtask}{subtask}
\newcommand{\subtasks}{subtasks}
\newcommand{\subspan}{subspan}

\title{\datasetname{}: A New Benchmark for Claim Detection from News\\ with Attribute Knowledge}

\author{Revanth Gangi Reddy$^1$, Sai Chetan Chinthakindi$^1$, Zhenhailong Wang$^1$, Yi R. Fung$^1$, \\ \textbf{Kathryn Conger$^2$, Ahmed Elsayed$^2$, Martha Palmer$^2$, Preslav Nakov$^3$,} \\ \textbf{Eduard Hovy$^{4,5}$, Kevin Small$^6$, Heng Ji$^1$} \\
$^1$UIUC \hspace{0.5em}  $^2$CU Boulder \hspace{0.5em} $^3$QCRI-HBKU \hspace{0.5em} $^4$DARPA \hspace{0.5em} $^5$CMU \hspace{0.5em} $^6$Amazon\\
  \texttt{\{revanth3,hengji\}@illinois.edu} \\
  }

\begin{document}
\maketitle
\begin{abstract}
Claim detection and verification are crucial for news understanding and have emerged as promising technologies for mitigating misinformation and disinformation in the news. However, most existing work has focused on {\em claim sentence} analysis while overlooking additional crucial attributes
%(e.g., claimer, claim objects)
(e.g., the claimer and the main object associated with the claim).
%, and other knowledge connected to the claim. 
In this work, we present \datasetname{}, a new benchmark for attribute-aware claim detection in the news domain. We extend the claim detection problem to include extraction of additional attributes related to each claim and release 889 claims annotated over 143 news articles. \datasetname{}\footnote{The code and data have been made publicly available here: \href{https://github.com/blender-nlp/NewsClaims}{https://github.com/blender-nlp/NewsClaims}} aims to benchmark claim detection systems in emerging scenarios, comprising unseen topics with little or no training data. %To this end, we provide a comprehensive evaluation of zero-shot and prompt-based baselines for \datasetname{}. 
To this end, we see that zero-shot and prompt-based baselines show promising performance on this benchmark, while still considerably behind human performance.
\end{abstract}

\input{sections/introduction}

\input{sections/related_work}

\input{sections/task}

\input{sections/dataset}

\input{sections/baselines}

\input{sections/experiments}

\section{Conclusion and Future Work}

We proposed a new benchmark, \datasetname{}, which extends the current claim detection task to extract more attributes related to each claim. Our benchmark comprehensively evaluates multiple aspects of claim detection such as identifying the topics, the stance, the claim span, the claim object, and the claimer in news articles from emerging scenarios such as the COVID-19 pandemic. We showed that zero-shot and prompt-based few-shot approaches can achieve promising performance in such low-resource scenarios, but still lag behind human performance, which presents opportunities for further research. In future work, we plan to explore extending this to build claim networks by identifying relations between the claims, including temporal connections. Another direction is build a unified framework that can extract claims and corresponding attributes together, without the need for separate components for each attribute.

\section*{Acknowledgement}

This research is based upon work supported by U.S. DARPA AIDA Program No. FA8750-18-2-0014.  The views and conclusions contained herein are those of the authors and should not be interpreted as necessarily representing the official policies, either expressed or implied, of DARPA, or the U.S. Government. The U.S. Government is authorized to reproduce and distribute reprints for governmental purposes notwithstanding any copyright annotation therein.

\section*{Limitations}
\datasetname{} exclusively consists of claims regarding COVID-19, which were intentionally chosen in order to sufficiently study a quickly emerging subject. However, the performance on this dataset might likely not be representative of the performance on a broader set of topics. \datasetname{} is not intended as a training dataset and a system using \datasetname{} in this way should be carefully evaluated before being used to annotate a larger dataset aimed at deriving journalism-centric conclusions. In the future, these risks can be mitigated by a larger dataset that can be more reliable to study these phenomena and to draw conclusions about the underlying media content.

\section*{Ethics and Broader Impact}

\paragraph{Annotator payment and approval} Our annotation process involved using both Turkers and expert annotators. For the first stage of annotation, Turkers were paid 15 cents per example (each example takes 30-35 seconds on average, meaning ~\$15 per hour). For the second stage, expert annotators were paid at an hourly rate, which was dependent on prior experience, but was always more than the usual rate of \$14 USD per hour. As per regulations set up by our organization’s IRB, this work was not considered to be human subjects research because no data or information about the annotators was collected, and thus it was IRB approval exempt.

\paragraph{Misuse Potential} The intended use of \datasetname{} is to evaluate methodological work regarding our augmented definition of claim detection, motivated by mitigating the spread of misinformation and disinformation in news media. However, given \datasetname{} is a smaller dataset over a set of hand-chosen topics, there is also potential for misuse. Specifically, \datasetname{} is not intended to directly make conclusions regarding the journalism quality nor quantify disagreement regarding the coverage of COVID-19 related topics.  As there has been continued controversy regarding media coverage of COVID-19, a bad faith or misinformed actor could produce artifacts that result in sensational, but potentially inaccurate, conclusions regarding COVID-19 claims in news media. 

%Do we want to say anything about actual data collection (i.e., trying to pay \$15/hr on average)?

\paragraph{Environmental Impact} We would also like to warn that the use of large-scale Transformers requires a lot of computations and the use of GPUs for training, which contributes to global warming \cite{strubell2019energy}. This is a bit less of an issue in our case, as we do not train such models from scratch; rather, we mainly use them in zero-shot and few-shot settings, and the ones we fine-tune are on relatively small datasets. All our experiments were run on a single 16GB V100.

\bibliography{anthology,custom}
\bibliographystyle{acl_natbib}

\clearpage
\appendix
\section{Appendix}
\label{sec:appendix}
\subsection{Annotation Interface}
\label{sec:app_annotation}
\begin{figure*}
    \centering
    \includegraphics[width=1\linewidth]{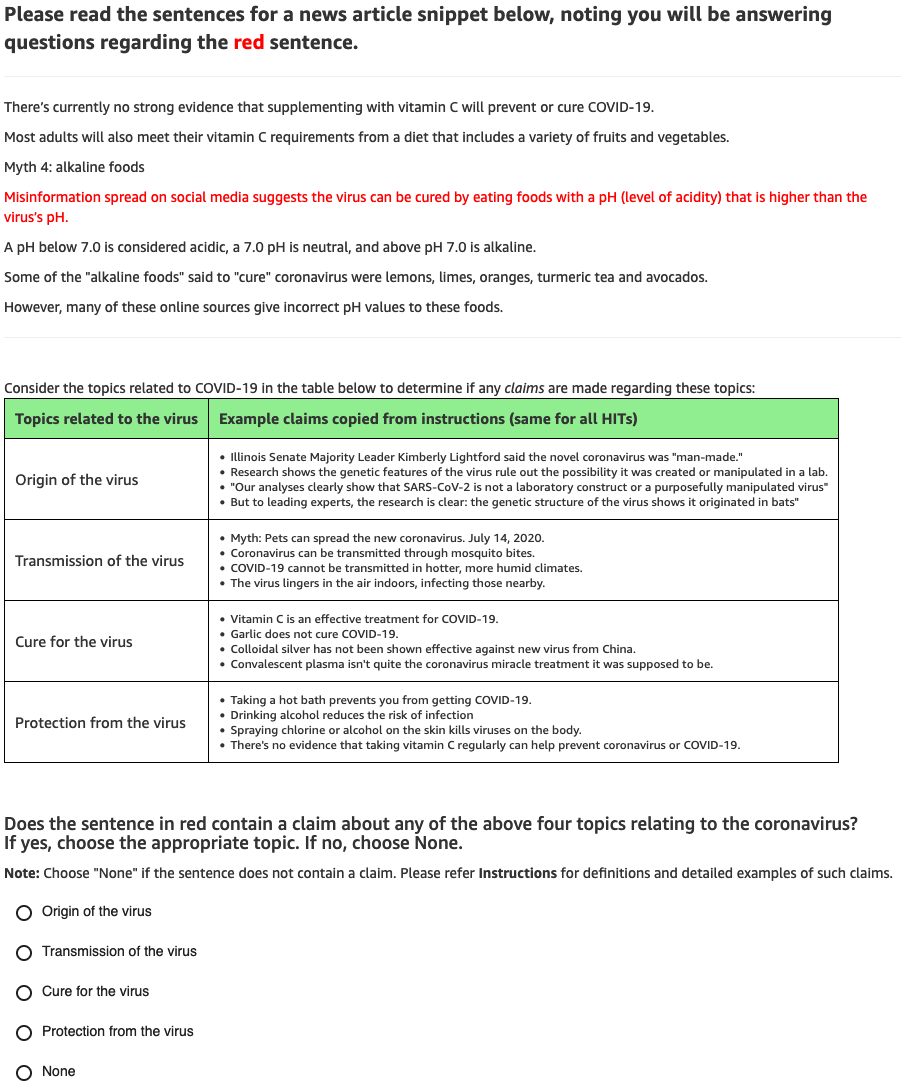}
    \caption{Screenshot of the phase 1 annotation interface.}
    \label{fig:interface_phase1}
\end{figure*}

\begin{figure*}
    \centering
    \includegraphics[width=1\linewidth]{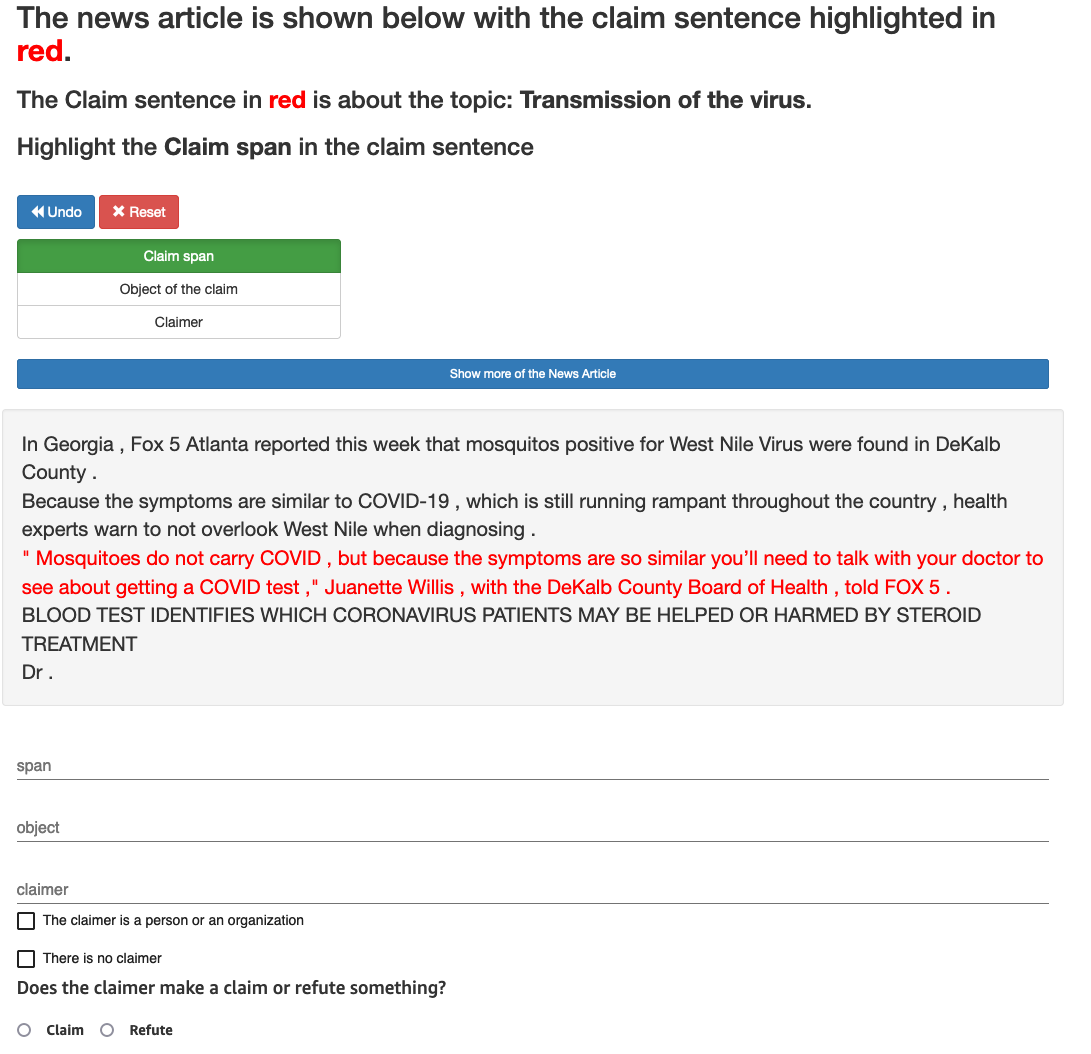}
    \caption{Screenshot of the phase 2 annotation interface.}
    \label{fig:interface_phase2}
\end{figure*}

In this section, we list the annotation guidelines and provide screenshots of the interface for both phases of annotation. Phase 1 of annotation involves identifying sentences which contain claims relating to a set of pre-defined topics about COVID-19. Phase 2 consists of annotating the attributes such as claimer, claimer's stance, claim object and the claimer span for each of the claims identified in phase 1. Figure \ref{fig:interface_phase1} and \ref{fig:interface_phase2} show screenshots of the annotation interface for phase 1 and 2 respectively. Below are some guidelines which we provide for detecting the claim sentences:

\begin{itemize}

    \item The highlighted sentence should be considered individually when deciding whether it contains a claim. The sentences around it are shown to provide context.
    
    \item Claims are usually statements made without presenting evidence or proof, and usually require further evidence to verify them. Sentences that just assert evidence or present facts should not be considered as claims.
    
    \item The claim sentences usually should also mention the object relating to the topic, i.e., which animal type  the virus came from, what conditions can transmit the virus, what can cure the virus or what can protect from the virus.
    
    \item Only those claims should be considered for which these topics can be directly inferred without any need for additional knowledge.
    
    \item Sentences that contain both claims as well as refute statements should be considered. For example, a  sentence that contains a statement that something cannot cure the coronavirus should be considered as containing a claim relating to the topic: Cure for the virus.
    
\end{itemize}

\subsection{SRL cue words}
\label{sec:app_srl}
Here, we list various cue words that we use to match against the verb predicates from the semantic role labeling system. These are categorized as affirming and refuting cue words, which are shown in tables \ref{tab:cue_affirm} and  \ref{tab:cue_refute} respectively.

\begin{table}[!htb]
    \centering
    \begin{tabular}{|c|}
\hline
accuse, affirm, allege, announce, argue \\
\hline
assert, aver, avouch, avow, blame\\
\hline
broadcast, claim, comment, confirm, contend\\
\hline
credit, declare, defend, describe, disclose\\
\hline
discuss, express, find, hint, imply\\
\hline
insinuate, insist, intimate, maintain, proclaim\\
\hline
profess, publish, purport, reaffirm, reassert\\
\hline
remark, repeat, report, restate, reveal\\
\hline
say, state, suggest, tell, write \\
\hline
    \end{tabular}
    \caption{Cue words corresponding to affirming a claim.}
    \label{tab:cue_affirm}
\end{table}

\begin{table}[!htb]
    \centering
    \begin{tabular}{|c|}
         \hline
challenge, controvert, contradict,disagree \\
\hline
discredit, dispute, deny, disavow, discount\\
\hline
protest, purport, reaffirm, question, repudidate\\
\hline
reject, repudiate, rebut, suppress, disaffirm \\
\hline
    \end{tabular}
    \caption{Cue words corresponding to refuting a claim.}
    \label{tab:cue_refute}
\end{table}

\subsection{GPT-3 prompt}
\label{sec:app_gpt3}

\begin{figure}[!htb]
    \centering
    \includegraphics[width=0.95\linewidth]{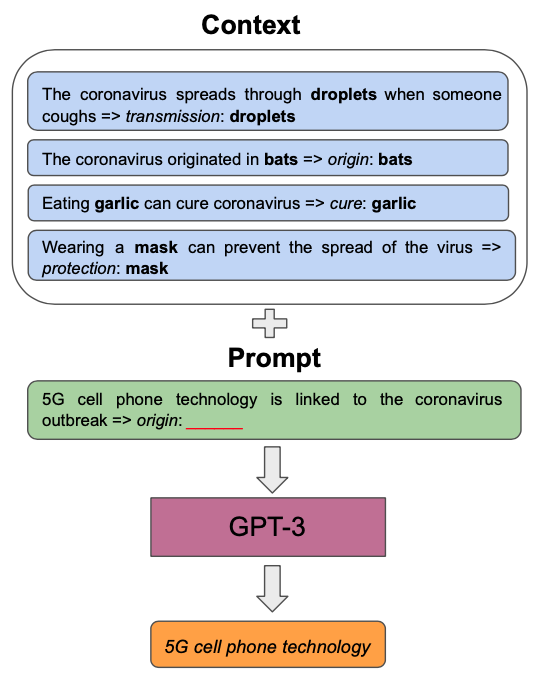}
    \caption{Figure showing the claim object detection sub-task input for GPT-3, with the few-shot labeled examples in context and the test example in the form of a prompt.}
    \label{fig:gpt3_prompt}
\end{figure}

In this section, we share more details of our approach for prompting GPT-3 for the claim object detection. In the in-context learning setting, we choose four examples from each topic as the few-shot examples. These labeled examples are then added to the context that is fed as input to GPT-3. The test example is added at the end of the context, in the form of a prompt, with the claim object to be generated by the system. Figure \ref{fig:gpt3_prompt} shows an example input along with the prompt.

\begin{figure*}
\centering
\begin{subfigure}[c]{0.47\linewidth}
    \centering
     \includegraphics[width=1\linewidth]{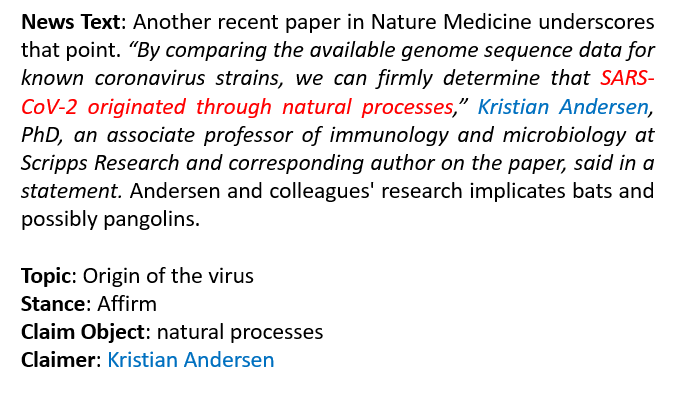}
       \label{fig:claim_freq}
\end{subfigure}
\hfill
\begin{subfigure}[c]{0.47\linewidth}
    \centering
     \includegraphics[width=1\linewidth]{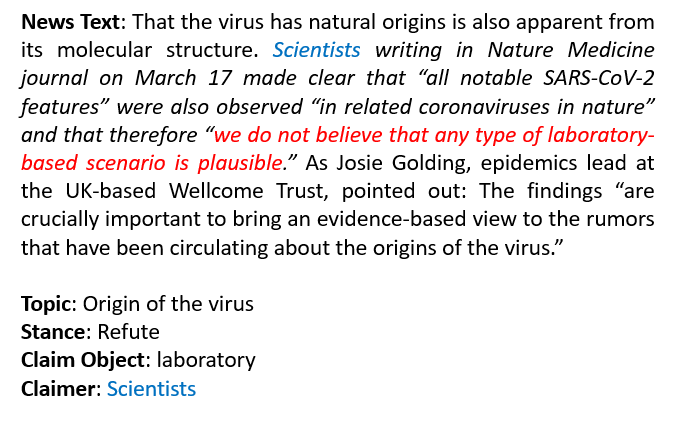}
       \label{fig:claim_freq}
\end{subfigure}
\begin{subfigure}[c]{0.47\linewidth}
    \centering
     \includegraphics[width=1\linewidth]{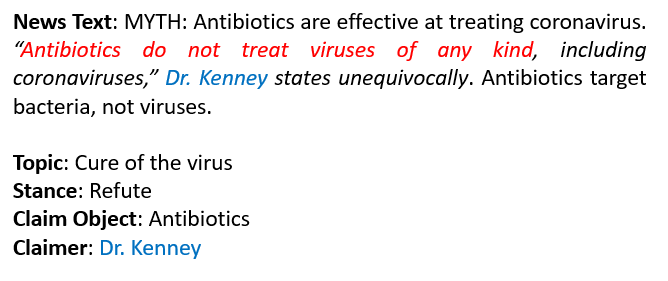}
       \label{fig:claim_freq}
\end{subfigure}
\hfill
\begin{subfigure}[c]{0.47\linewidth}
    \centering
     \includegraphics[width=1\linewidth]{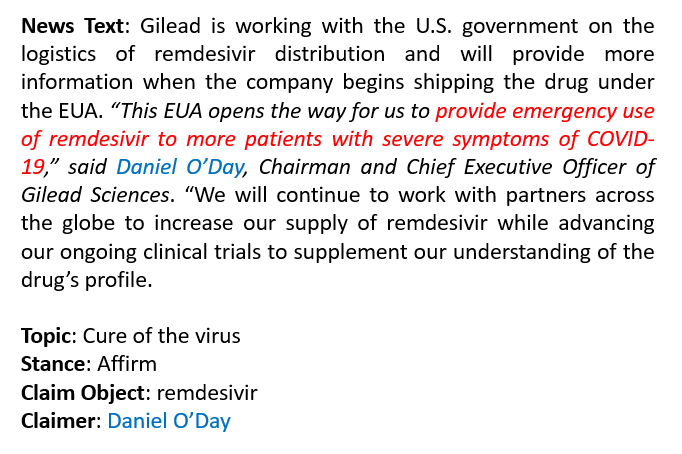}
       \label{fig:claim_freq}
\end{subfigure}
\begin{subfigure}[c]{0.47\linewidth}
    \centering
     \includegraphics[width=1\linewidth]{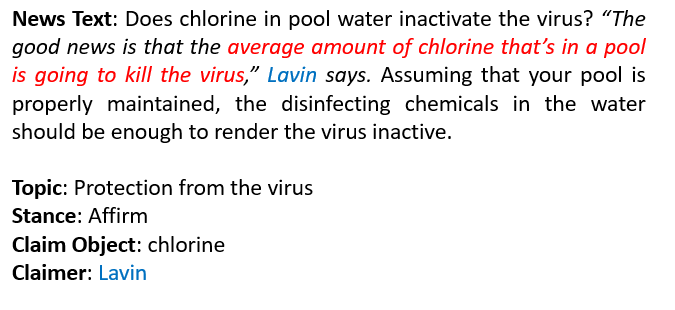}
       \label{fig:claim_freq}
\end{subfigure}
\hfill
\begin{subfigure}[c]{0.47\linewidth}
    \centering
     \includegraphics[width=1\linewidth]{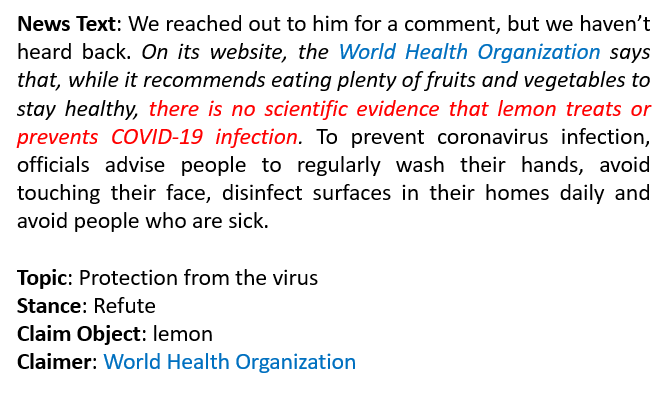}
       \label{fig:claim_freq}
\end{subfigure}
\begin{subfigure}[c]{0.47\linewidth}
    \centering
     \includegraphics[width=1\linewidth]{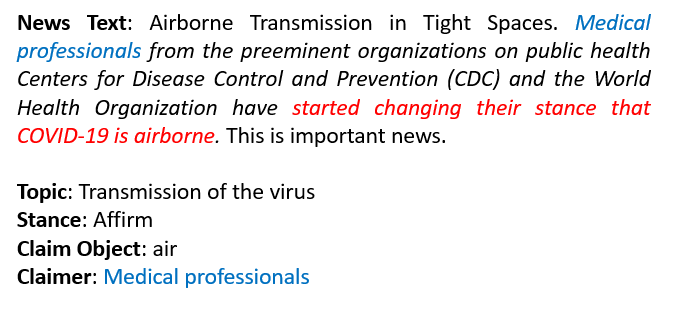}
       \label{fig:claim_freq}
\end{subfigure}
\hfill
\begin{subfigure}[c]{0.47\linewidth}
    \centering
     \includegraphics[width=1\linewidth]{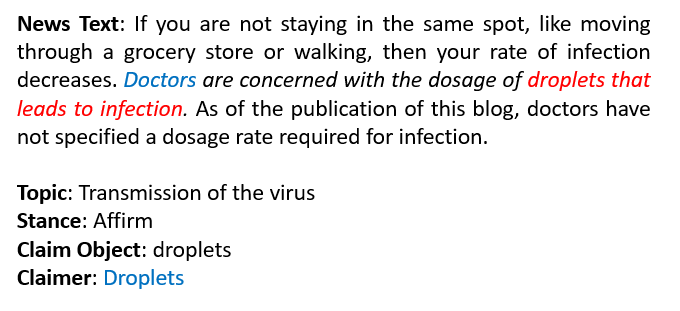}
       \label{fig:claim_freq}
\end{subfigure}
\caption{Some examples from the \textsc{NewsClaims} benchmark.}
\end{figure*}

\end{document}

%% file: sections/introduction.tex
\section{Introduction}

The internet era has ushered in an explosion of online content creation, resulting in increased concerns regarding misinformation in news, online debates, and social media. A key element of identifying misinformation is detecting the claims and the arguments that have been presented. In this regard, news articles are particularly interesting as they contain claims in various formats: from arguments by journalists to reported statements by prominent public figures.

Check-worthiness estimation aims to decide if a piece of text is worth fact-checking, i.e., whether it contains an important verifiable factual claim~\cite{hassan2017toward}. Most current approaches \cite{jaradat-etal-2018-claimrank, shaar2021overview} largely ignore relevant attributes of the claim (e.g.,~the claimer and the primary object associated with the claim). Moreover, current claim detection tasks mainly identify claims in debates \cite{gencheva-etal-2017-context}, speeches \cite{atanasova2019overview}, and social media \cite{CLEF2022:CheckThat:ECIR}, where the claim source (i.e.,~the claimer) is known.

%as they have  mainly dealt with data written in a collective style (e.g., Wikipedia) or from a monotone viewpoint \cite{stab2014identifying} (e.g., persuasive essays). 
%\heng{Here we should elaborate the unique challenges of doing claim detection from news, compared to Wikipedia or social media} %In such data domains, the reading focus sheds on details inside the claims and premises. 

\begin{figure}[t]
    \centering
    \includegraphics[scale=0.21]{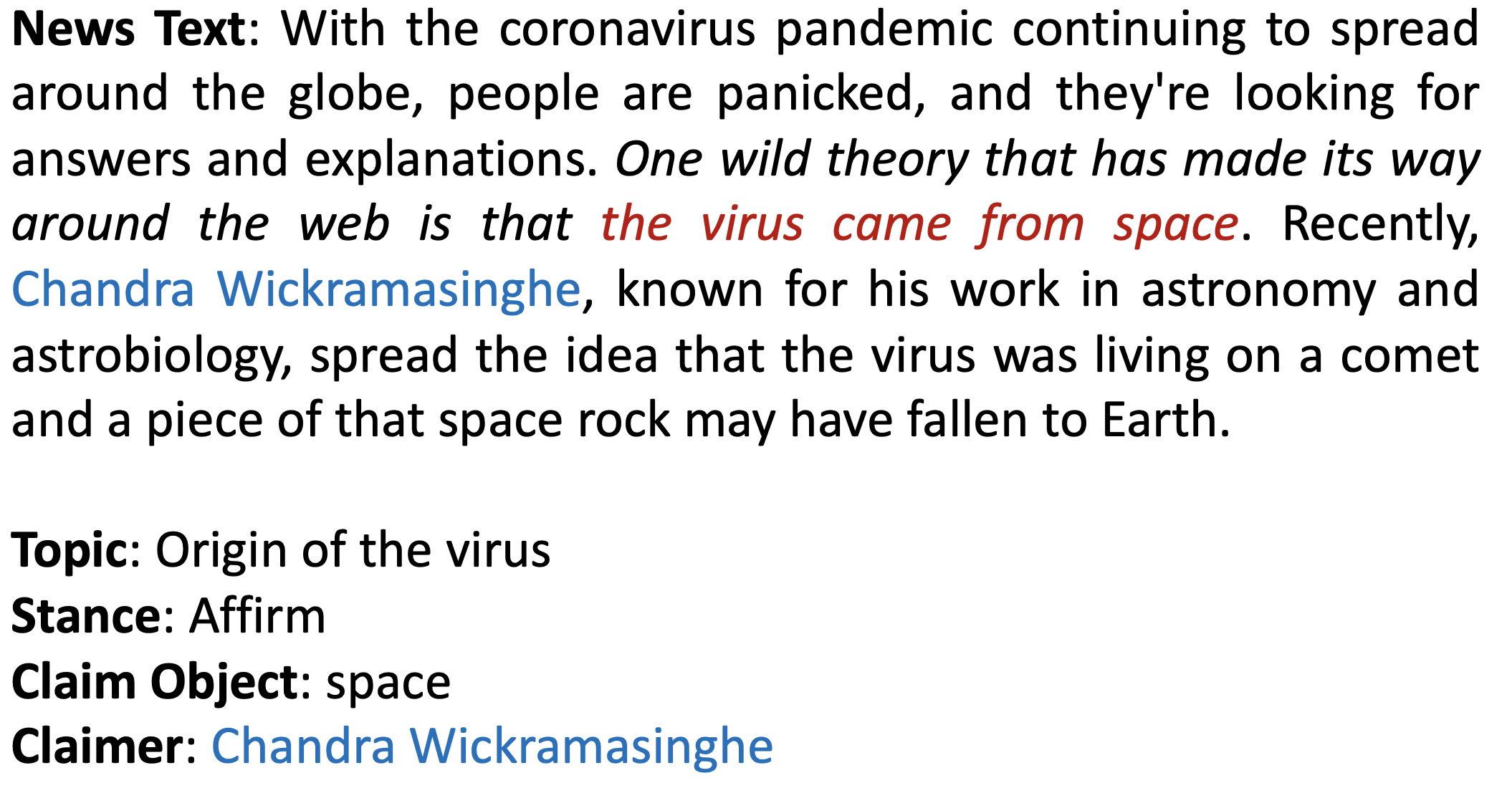}
    \caption{A news article containing a claim regarding the origin of COVID-19 with the \emph{claim sentence} in italics, the \emph{claim span} in red, and the \emph{claimer} in blue. Also shown are the \emph{claimer stance} and the \emph{claim object}.}
    \label{fig:intro_ex}
\end{figure}

%However, 
%information domains such as 
News articles, on the other hand, have more complex arguments, requiring a deeper understanding of what each claim is about and identifying where it comes from.
Thus, here we introduce the notion of \emph{claim object}, which we define as an entity that identifies what is being claimed with respect to the topic of the claim. Figure~\ref{fig:intro_ex} shows a claim about the origin of COVID-19, suggesting that the virus came from \emph{space}, which is the claim object. 
We further identify the \emph{claimer}, which could be useful for fact-checking organizations to examine how current claims compare to previous ones by the same person/organization.
In this regard, we extend the claim detection task to ask for the extraction of more attributes related to the claim. Specifically, given a news article, we aim to extract all \emph{claims} pertaining to a set of \emph{topics} along with the corresponding \emph{claim span}, the \emph{claimer}, the \emph{claimer's stance}, and the \emph{claim object} for each claim. The claim attributes enable comparing claims at a more fine-grained level: claims with the same topic, object and stance can be considered equivalent whereas those with similar claim objects but opposing stance could be contradicting. 
%Moreover, all claims from the same source can be summarized to present the overall opinion %of an entity and to improve news comprehension.
 We note that while identifying the claim span and stance have been explored independently in prior work \cite{levy2014context, hardalov-etal-2021-cross}, we bring them into the purview of a unified claim detection task.

To promote research in this direction, we release \datasetname{}, a new evaluation benchmark for claim detection.
%with 889 claims annotated over 143 news articles.
We consider this in an evaluation setting since 
%harmful content\footnote{\href{https://ai.facebook.com/blog/harmful-content-can-evolve-quickly-our-new-ai-system-adapts-to-tackle-it}{harmful-content-blog-post}} can evolve rapidly
claims about new topics can emerge rapidly\footnote{\href{https://ai.facebook.com/blog/harmful-content-can-evolve-quickly-our-new-ai-system-adapts-to-tackle-it}{harmful-content-blog-post}}, requiring systems that are effective under zero/few-shot settings. \datasetname{} aims to study how existing NLP techniques can be leveraged to tackle claim detection in emerging scenarios and regarding previously unseen topics. We explore multiple zero/few-shot strategies for our \subtasks{} including topic classification, stance detection, and claim object detection. This is in line with recent progress in using pre-trained language models in zero/few-shot settings \cite{brown2020language, liu2021pre}. Such approaches can be adapted to new use cases and problems as they arise without the need for large additional training data.
%over multiple topics relating to the coronavirus. In total, the evaluation benchmark contains X claims over Y news articles related to the COVID-19 pandemic. 
%Figure~\ref{fig:intro_ex} shows an example from \datasetname{}, including an extracted claim and its background attributes.

In our benchmark, all news articles are related to the COVID-19 pandemic, motivated by multiple considerations. First, COVID-19 has gained extensive media coverage, with the World Health Organization coining the term \emph{infodemic}\footnote{\href{https://www.un.org/en/un-coronavirus-communications-team/un-tackling-\%E2\%80\%98infodemic\%E2\%80\%99-misinformation-and-cybercrime-covid-19}{COVID-19 Infodemic}} to refer to disinformation related to COVID-19 \cite{naeem2020covid} and suggesting that ``fake news spreads faster and more easily than this virus''. Second, this is an emerging scenario with limited previous data related to the virus, making it a suitable candidate for evaluating claim detection in a low-resource setting. \datasetname{} covers claims about four COVID-19 topics, namely the origin of the virus, possible cure for the virus, the transmission of the virus, and protecting against the virus.

Our contributions include %\heng{Would you like to add some deep analysis about remaining challenges? maybe it will be good to line up some research directions to tackle these challenges. I think that will be a nice contribution.}
%\begin{itemize}
    (\emph{i})~extending the claim detection task to include more attributes (claimer and object of the claim),
    (\emph{ii})~releasing a manually annotated evaluation benchmark for this new task, \datasetname{}, which covers multiple topics related to COVID-19 
%    To the best of our knowledge, 
     and is the first dataset with such extensive annotations for claim detection in the news, with 889 claims from 143 news articles, and
    % \heng{summarize some data stats here}
    %(\emph{iii})~performing comprehensive evaluation of multiple zero-shot and prompt-based few-shot approaches for the components of our claim detection task.
    (\emph{iii})~demonstrating promising performance of various zero-shot and prompt-based few-shot approaches for the claim detection task.

    %\item We show that a question answering system can be effectively leveraged for multiple aspects of the claim detection task, without needing any task-specific training data.
%\end{itemize}

%In this paper, we point out the importance of outer frame claim attributes and present an unified framework to detect the inner and outer frames of claims together. Specifically, we observe that it may be computationally redundant to build a different model to detect each outer frame attribute separately from the inner claim. We propose to: 1) leverage multitask learning for joint inner-outer frame claim detection through an architecture outlined in fig \ref{fig:my_label}, 2) use knowledge elements to improve claim detection, and 3) utilize a question answering (QA) framework for outer frame attribution in the scenario when annotation is lacking.

%% file: sections/related_work.tex
\section{Related Work}

\begin{figure*}
    \centering
    \includegraphics[scale=0.36]{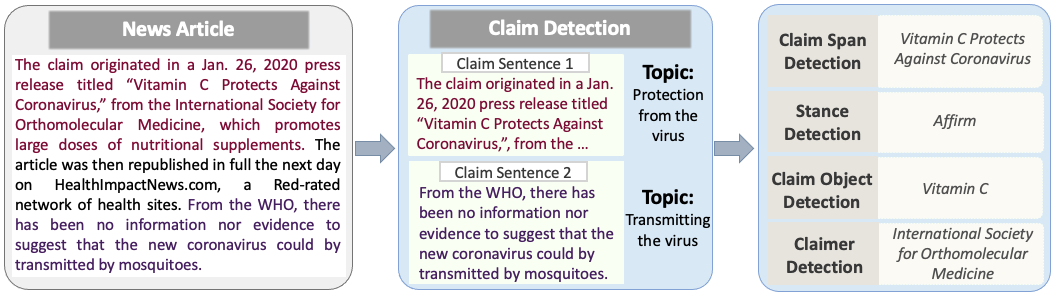}
    \caption{An example demonstrating our proposed claim detection task, and its \subtasks{}. The following attributes are to be extracted for each claim: the \emph{claimer}, \emph{claimer's stance}, \emph{claim object}, and \emph{claim span}.}
    \label{fig:ex_pipeline}
\end{figure*}

%\begin{figure*}
%    \centering
%    \includegraphics[scale=0.5,trim={0cm 166 0 130}]{images/ClaimDetection_Pipeline_Yi.pdf}
%    \caption{An example demonstrating our proposed claim detection task, and its sub-tasks. The background attributes such as the claim span, its stance, its object, and its claimer need to be identified for each of the claims in the news article.}
%    \label{fig:ex_pipeline}
%\end{figure*}
%\label{sec:task}
Automatic fact-checking has a number of sub-tasks such as detecting check-worthy claims \cite{jaradat-etal-2018-claimrank,vasileva2019takes}, comparing them against previously-fact checked claims \cite{shaar2020known,nakov2021automated}, retrieving evidence relevant to a claim \cite{karadzhov2017fully,augenstein2019multifc} and finally inferring the veracity of the claim \cite{karadzhov2017fully,thorne2018fever,atanasova2019automatic}. Our work here is positioned in the space of identifying check-worthy claims, also known as \emph{check-worthiness estimation}. In this work, we show that identifying the topic of the claim is beneficial, by leveraging it towards stance detection (Section~\ref{sec:stance_det_baseline}) and claim object detection (Section~\ref{sec:claim_object_baseline}).

Argumentation mining~\cite{palau2009argumentation,stab2014identifying,stab2018cross} includes context-dependent claim detection \cite{levy2014context,levy2017unsupervised}, which entails detecting claims specifically relevant to a predefined topic. However, claims in the context of argumentation are neither necessarily factual nor verifiable.
%\emph{Argument mining} \cite{palau2009argumentation} involves automatically detecting arguments in text. Structured argumentation mining \cite{stab2014identifying, eger2017neural} aims to identify argument components and relations with respect to each other, which differs from the general analysis of opinions in unstructured argumentation mining \cite{stab2018cross}. On the other hand, context-dependent claim detection \cite{levy2014context} involves detecting claims specifically relevant to a predefined topic, whereas \citet{lippi2015context} proposed a context-independent claim detection task in which one attempts to detect claims without a specified input topic. Corpus-wide claim detection \cite{levy2017unsupervised} aims to mine arguments from large text corpora to build an argumentative content search engine \cite{levy2018towards}.
Moreover, prior work on both check-worthiness estimation and argumentation mining did not deal with identifying additional claim attributes, such as the claimer, or the source of the claim, and the claim object.

The claimer detection \subtask{} is related to attribution in the news. Current attribution methods are mainly sentence-level \cite{pareti2016parc} or only involve direct quotations \cite{elson2010automatic}. In contrast, we require cross-sentence reasoning for identifying the claimer as it may not be present in the claim sentence (see Figure~\ref{fig:intro_ex}).

There has been recent work addressing claims related to COVID-19. \citet{saakyan-etal-2021-covid} proposed a new FEVER-like \cite{thorne2018fever} dataset, where given a claim, the task is to identify relevant evidence and to verify whether it refutes or supports the claim; however, this does not tackle identifying the claims or the claimer. There has also been work on identifying the check-worthiness of tweets related to COVID-19 \cite{alam2020fighting, jiang2021categorising}; however, unlike news articles, tweets do not require attribution for claimer identification.

%In contrast, our benchmark requires document-level inference for identifying the claimer, which need not be present in the claim sentence (see example in Figure \ref{fig:intro_ex}).

%Recent work \cite{du2020event, liu2020event, zhou2021role} has leveraged question answering systems for the task of event argument extraction, by fine-tuning with event argument data. However, we apply the QA system for claim attribute extraction in a strict zero-shot setting. In addition, we leverage the same QA system for more tasks like topic and claim boundary detection.

%% file: sections/task.tex
\section{Proposed Claim Detection Task}
\label{sec:task}
%The \datasetname{} 
Our task is to identify claims related to a set of topics in a news article along with corresponding attributes such as the claimer, the claim object, and the claim span and stance, as shown in Figure~\ref{fig:ex_pipeline}.
%In this section, we describe these \subtasks{} in detail.
%Figure~\ref{fig:ex_pipeline} shows the expected output from each \subtask{} for a given input news article.

\noindent \textbf{Claim Sentence Detection}:
Given a news article, the first \subtask{} is to extract claim sentences relevant to a set of pre-defined topics. This involves first identifying sentences that contain \textit{factually verifiable claims}, similar to prior work on check-worthiness estimation, and then selecting %filtering out 
those that are related to the target topics. To address misinformation in an emerging real-world setting, we consider the following topics related to COVID-19:
%\footnote{Sample claims for each topic are shown in the appendix.}:
%\begin{itemize}
    \textit{\textbf{Origin of the virus}}: claims related to the origin of the virus (i.e., location of first detection, zoonosis, `lab leak' theories);
    \textit{\textbf{Transmission of the virus}}: claims related to who/what can transmit the virus or conditions favorable for viral transmission;
    \textit{\textbf{Cure for the virus}}: claims related to curing the virus, (e.g.,~via medical intervention after infection); and
    \textit{\textbf{Protection from the virus}}: claims related to precautions against viral infection.
%\end{itemize}

\noindent \textbf{Claimer Detection}: Claims within a news article can come from various types of sources such as an entity (e.g., person, organization) or published artifact (e.g., study, report, investigation). In such cases, the claimer identity can usually be extracted from the news article itself. However, if the claim is asserted by the article author or if no attribution is specified or inferrable, then the article author, i.e. the journalist, is considered to be the claimer. The claimer detection \subtask{} involves identifying whether the claim is made by a \emph{journalist} or whether it is \emph{reported}\label{sec:task} in the news article, in which case the source is also extracted. Moreover, sources of such reported claims need not be within the claim sentence. In our datatset \datasetname{}, the claimer span was extracted from outside of the claim sentence for about 47\% of the claims. Thus, the claimer detection \subtask{} in our benchmark requires considerable document-level reasoning, thus making it harder than existing attribution tasks \cite{pareti-2016-parc, newell2018attribution}, which require only sentence-level reasoning.

\noindent \textbf{Claim Object Detection}: The claim object relates to what is being claimed in the claim sentence with respect to the topic. For example, in a claim regarding the virus origin, the claim object could be the species of origin in zoonosis claims, or who created the virus in bioengineering claims. Table~\ref{tab:ex_claim_object} shows examples of claim objects %for example claim sentences 
from each topic. We see that the claim object is usually an extractive span within the claim sentence. 
Identifying the claim object helps to better understand the claims and potentially identify claim--claim relations, since two claims with the same object are likely to be similar.

\begin{table}[!htb]
\small   
%\scriptsize
\centering
    \begin{tabular}{|l|p{16em}|}
    %\begin{tabular}{|p{9.5em}|c|c|}
       \hline
       \textbf{Topic} & \textbf{Claim Sentence}  \\
       \hline
       Origin & The genetic data is pointing to this virus coming from a \textbf{bat reservoir}, he said.\\
       \hline
       Transmission & The virus lingers in the \textbf{air indoors}, infecting those nearby \\
       \hline
       Cure & \textbf{Vitamin C} is an effective treatment for COVID-19. \\
       \hline
       Protection & Taking a \textbf{hot bath} prevents you from getting COVID-19.  \\
       \hline
    \end{tabular}
    \caption{Examples showing the claim object in \textbf{bold} for claims corresponding to \datasetname{} topics.}
    \label{tab:ex_claim_object}
\end{table}

\begin{figure*}[!htb]
     \begin{subfigure}[c]{0.35\linewidth}
    \centering
     \includegraphics[width=1\linewidth]{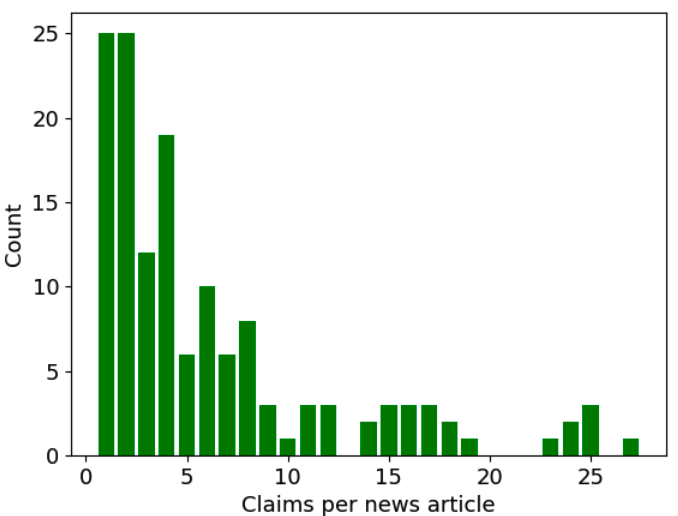}
       \caption{claim counts per news article}
       \label{fig:claim_counts}
     \end{subfigure}
     \begin{subfigure}[c]{0.31\linewidth}
     \centering
     \includegraphics[width=1\linewidth]{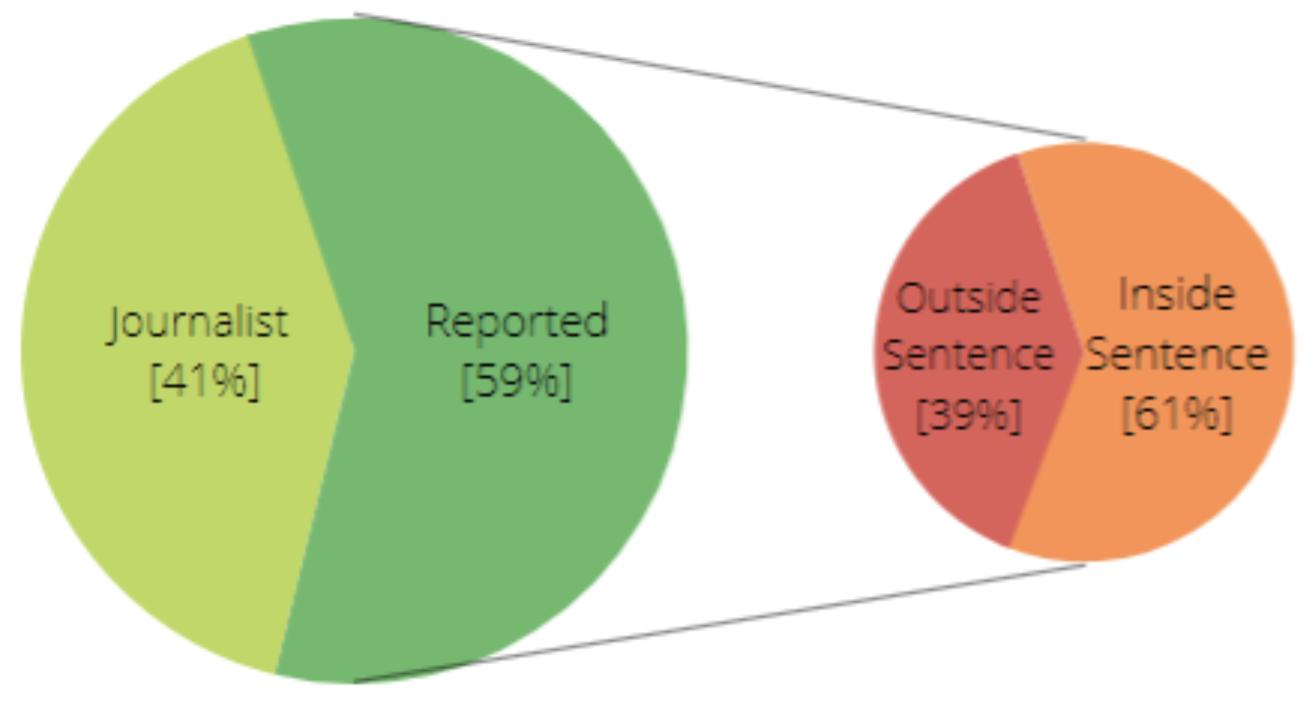}
       \caption{claims by journalists vs. reported ones, along with claimer coverage for reported claims\vspace{-2.3em}}
       \label{fig:claimer_dist}
     \end{subfigure}
     \hfill
      \begin{subfigure}[c]{0.31\linewidth}
     \centering
     \includegraphics[width=1\linewidth]{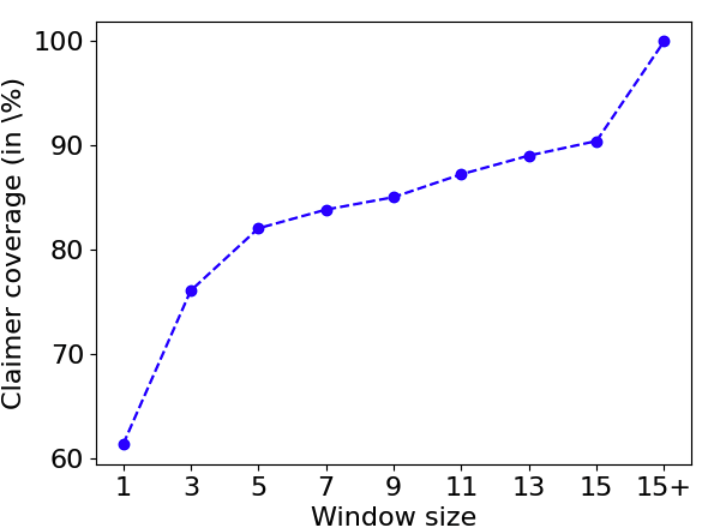}
       \caption{coverage of claimer within a window size based on number of sentences around the claim sentence\vspace{-0.2em}}
       \label{fig:claimer_coverage}
     \end{subfigure}
     \hfill
\caption{\datasetname{} benchmark statistics: (a) number of claims per news article, (b) claims by journalists vs. reported claims, and (c) claimer coverage by window size within the news article for reported claims.}
     \label{fig:dataset_stats}
\end{figure*}

\noindent \textbf{Stance Detection}: This \subtask{} involves outputting whether the claimer is asserting (\textit{affirm}) or refuting (\textit{refute}) a claim within the given claim sentence. We note that stance detection in \datasetname{} differs from the task formulation used in other stance detection datasets \cite{stab2018cross, hanselowski2019richly, allaway-mckeown-2020-zero} as it involves identifying the claimer's stance within a claim sentence -- whereas prior stance detection tasks, as described in a recent survey by \citet{hardalov2021survey}, involve identifying the stance for \textit{target--context} pairs. For example, given pairs such as claim--evidence or headline--article, it involves identifying whether the evidence/article at hand supports or refutes a given claim/headline.

\noindent \textbf{Claim Span Detection}: Given a claim sentence, this \subtask{} aims to identify the exact claim boundaries within the sentence, including the actual claim content, usually without any cue words (e.g.,~\emph{asserted}, \emph{suggested}) and frequently a contiguous \subspan{} of the claim sentence. Identifying the precise claim conveyed within the sentence can be useful for downstream tasks such as clustering claims and identifying similar or opposing claims.

%\textbf{TODO: Revanth}:  Add some justification on why claimer extraction is important

%% file: sections/dataset.tex
\section{The \datasetname{} Dataset}

In this work, we build \datasetname{}, a new benchmark dataset for evaluating the performance of models on different components of our claim detection task. Specifically, we release an evaluation set based on news articles about COVID-19, which can be used to benchmark systems on detecting claim sentences and associated attributes including claim objects, claim span, claimer, and claimer stance. 
\datasetname{} uses news articles from the LDC corpus \textit{LDC2021E11}, from which we selected those related to COVID-19. We describe below the annotation process (Section~\ref{sec:annotation}) and provide statistics about \datasetname{} (Section~\ref{sec:statistics}).

%\heng{If you end up with adding documents from Zhenhailong, make it clear that they are from the news documents in the reference sections of the Wikipedia pages related to COVID}

\subsection{Annotation}
\label{sec:annotation}
Given a news article, we split the annotation process into two phases: (\emph{i})~identifying claim sentences with their corresponding topics, and (\emph{ii})~annotating the attributes for these claims.\footnote{Detailed annotation guidelines and screenshots of the interface are provided in Section~\ref{sec:app_annotation} in the appendix.}
In the first phase, the interface displays the entire news article with a target sentence highlighted in red. The annotators are asked whether the highlighted sentence contains a claim associated with the four pre-defined COVID-19 topics and to indicate the specific topic if that is the case. 
In the second phase, the interface displays the entire news article with a claim sentence highlighted in red. The annotators are asked to identify the claim span, the claim object, and the claimer from the news article. The annotators are also asked to indicate the claimer's stance regarding the claim. We provide a checkbox to use if there is no specified claimer, in which case the journalist is considered to be the claimer. 

%\begin{figure*}[!htb]
%     \begin{subfigure}[c]{0.40\linewidth}
%    \centering
%     \includegraphics[width=1\linewidth]{images/dist_hist.png}
%       \caption{number of claims per news article}
%       \label{fig:claim_freq}
%     \end{subfigure}
%     \begin{subfigure}[c]{0.25\linewidth}
%     \centering
%     \includegraphics[width=1\linewidth]{images/topic_dist_2.png}
%       \caption{topics\vspace{-1.7em}}
%       \label{fig:topic_dist}
%     \end{subfigure}
%     \hfill
%     \begin{subfigure}[c]{0.31\linewidth}
%     \centering
%     \includegraphics[width=1\linewidth]{images/claimer_dist_2.png}
%       \caption{claimer by source and presence in the news article\vspace{-4.0em}}
%       \label{fig:claimer_dist}
%     \end{subfigure}
%\caption{Statistics about our dataset.}
%     \label{fig:dataset_stats}
%\end{figure*}
For the first stage of annotation, which involves identifying claim sentences (and their topics) from the entire news corpus, we used 3 annotators per example hired via Mechanical Turk \cite{buhrmester2011amazon}. Only sentences with unanimous support were retained as valid claims. For the second stage, which involves identifying the remaining attributes (claim object, span, claimer, and stance), we used expert annotators to ensure quality, with 1 annotator per claim sentence.
%We used Amazon Mechanical Turk \cite{buhrmester2011amazon} for annotation, assigning three annotators per example. In phase one, only sentences with unanimous support were retained as valid claims. In the second phase, majority voting was used to determine the stance and the claimer. 
Annotators took ${\sim}30$ seconds per sentence in the first phase and ${\sim}90$ seconds to annotate the attributes of a claim in phase two. For claim sentence detection, the inter-annotator agreement had a Krippendorff's kappa of 0.405, which is moderate agreement; this is on par with previous datasets that tackled identifying topic-dependent claims \cite{kotonya2020explainable, bar2020arguments}, which is more challenging than topic-independent claim annotation \cite{thorne2018fever, aly2021feverous}.

\subsection{Statistics}

\label{sec:statistics}
\datasetname{} consists of development and test sets with 18 articles containing 103 claims and 125 articles containing 786 claims, respectively. The development set can be used for few-shot learning or for fine-tuning model hyper-parameters. Figure~\ref{fig:claim_counts} shows a histogram of the number of claims in a news article where most news articles contain up to 5 claims, but some have more than 10 claims. Claims related to the origin of the virus are most prevalent, with the respective topic distribution being 35\% for origin, 22\% for cure, 23\% for protection, and 20\% for transmission.  Figure~\ref{fig:claimer_dist} shows the distribution of claims by journalists vs. reported claims: we can see that 41\% of the claims are made by journalists, with the remaining 59\% coming from sources mentioned in the news article. Moreover, for reported claims, the claimer is present outside of the claim sentence 39\% of the time, demonstrating the document-level nature of this task. Figure~\ref{fig:claimer_coverage} shows the claimer coverage (in \%) based on a window around the claim by the number of sentences and indicates that document-level reasoning is required to identify the claimer, with some cases even requiring inference beyond a window size of 15. Note that the 61\% inside-sentence coverage in Figure~\ref{fig:claimer_dist} corresponds to a window size of 1 in Figure~\ref{fig:claimer_coverage}.

%% file: sections/baselines.tex
\section{Baselines}

\begin{figure*}[ht]
     \begin{subfigure}[b]{0.48\textwidth}       
     \includegraphics[scale=0.12]{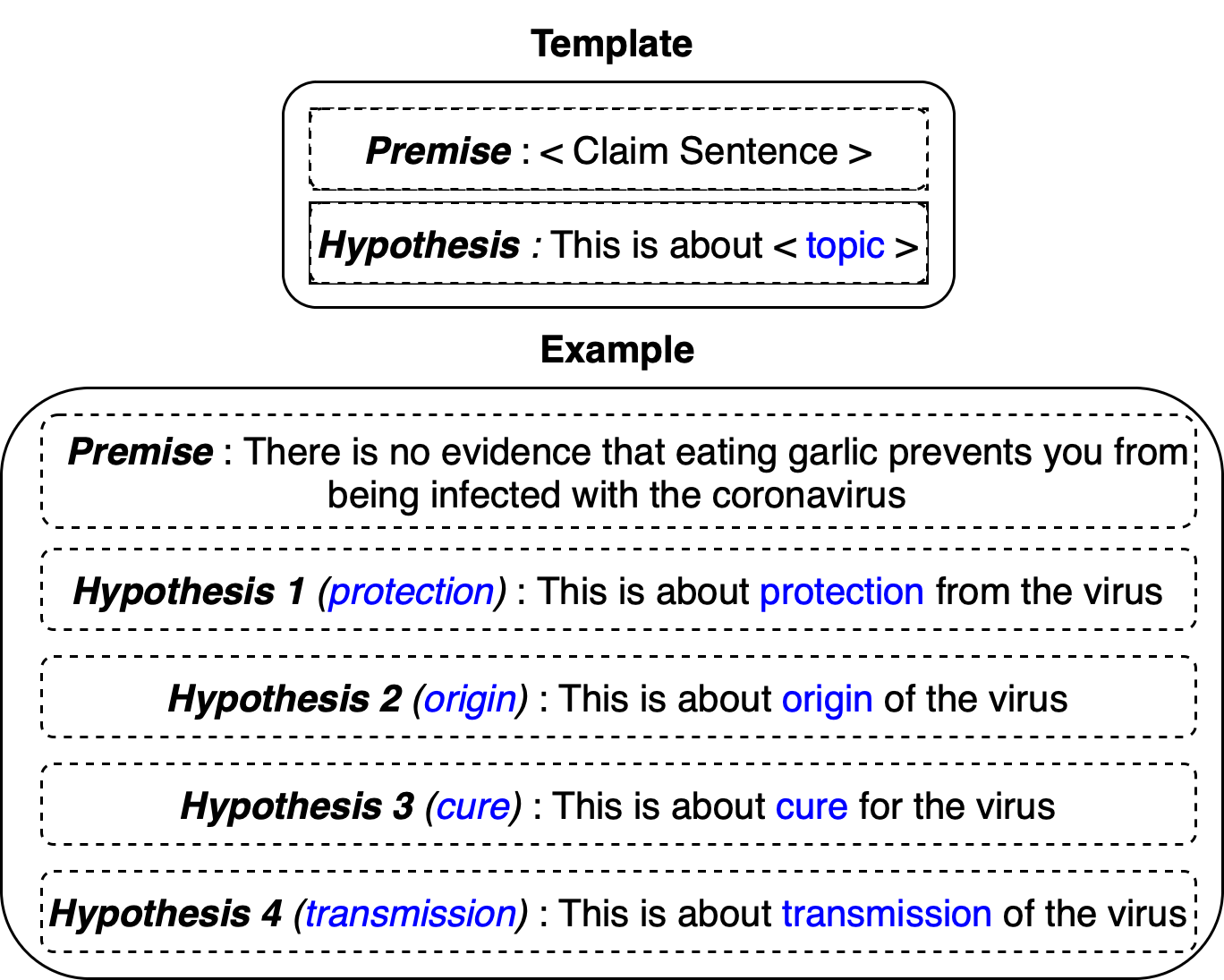}       \caption{zero-shot NLI for topic classification}
       \label{fig:mnli_topic_detection}
     \end{subfigure}
     \hfill
     \begin{subfigure}[b]{0.51\textwidth}       
     \includegraphics[scale=0.13]{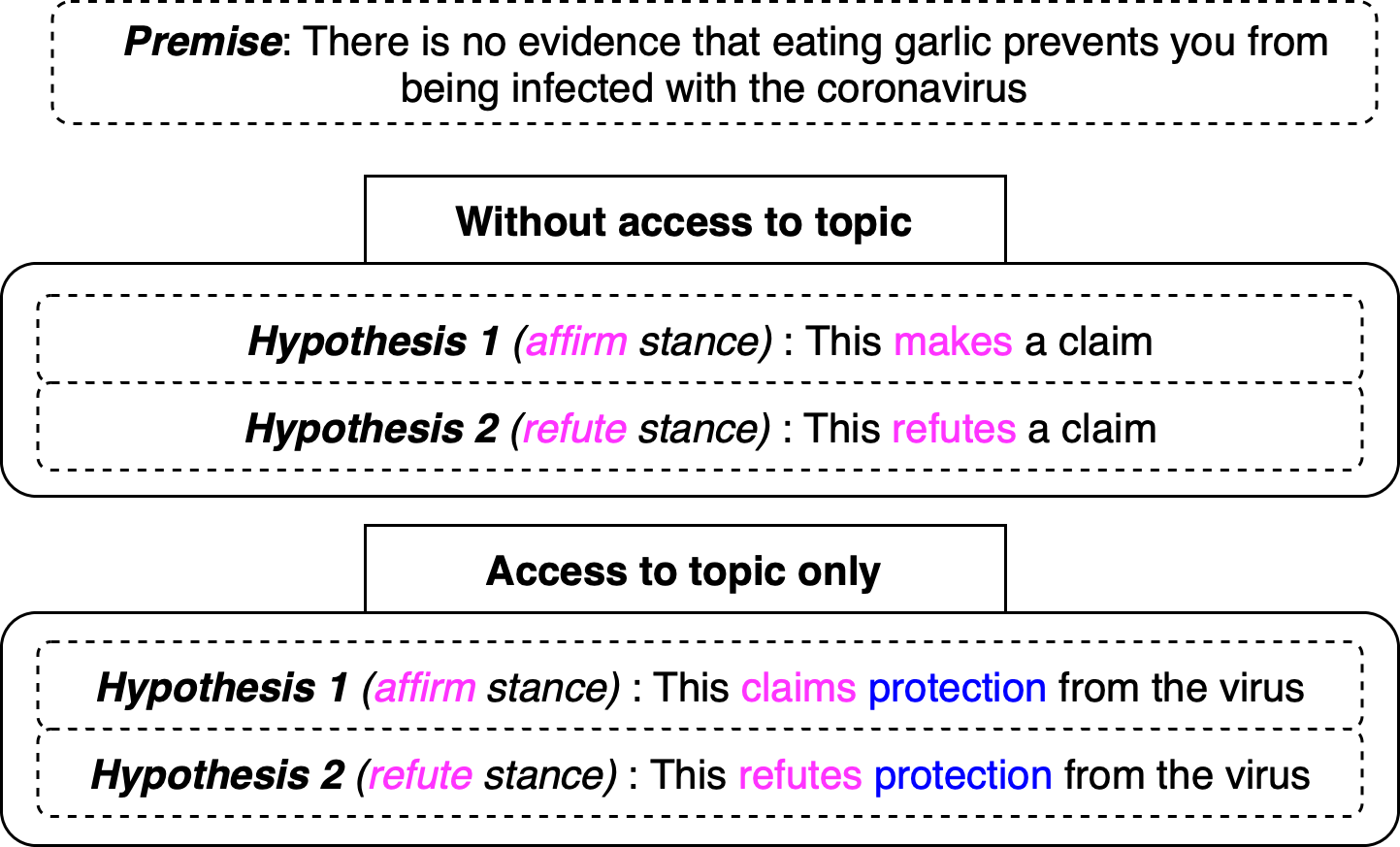}       \caption{zero-shot NLI for stance detection}
       \label{fig:mnli_stance_detection}
     \end{subfigure}
     \caption{Diagram (a) shows the template and an example for leveraging a pre-trained NLI model for zero-shot topic classification; topic corresponding to the hypothesis with the highest entailment score is taken as the claim sentence topic. Diagram (b) shows examples for leveraging a pre-trained NLI model for zero-shot stance detection. Each example shows hypothesis construction based on the class label (in pink) and the topic (in blue).}
     \label{fig:MNLI}
\end{figure*}

In this section, we describe various zero-shot and prompt-based few-shot learning baselines for the claim detection \subtasks{} outlined in Section~\ref{sec:task}. We describe a diverse set of baselines with each chosen to be relevant in an evaluation-only setting.

\subsection{Claim Sentence Detection}
\label{sec:claim_det_baseline}

Given a news article, we aim to detect all sentences that contain claims related to a pre-defined set of topics regarding COVID-19. We use a two-step procedure that first identifies sentences that contain claims and then selects those related to COVID-19.

\paragraph{Step 1. ClaimBuster:} To identify sentences containing claims, we use ClaimBuster \cite{hassan2017claimbuster},\footnote{\href{https://idir.uta.edu/claimbuster/api/}{https://idir.uta.edu/claimbuster/api/}} a claim-spotting system trained on a dataset of check-worthy claims~\cite{arslan2020benchmark}.
%which outputs a check-worthiness score for each sentence in the input news article, which we use to select sentences that contain claims. 
As ClaimBuster has no knowledge about topics, we use zero-shot topic classification, as described below. 
%, as described next, for filtering out claims that are related to topics about the coronavirus. For the claim sentences output by the ClaimBuster model, the scores for each topic are computed and the highest score is used while filtering.\\

 %We leverage additional zero-shot models, as we describe next, for filtering out claims that are related to topics about the coronavirus. For the claim sentences output by the ClaimBuster model, the scores for each topic are computed and the highest score is used while filtering. \\

%\noindent \textbf{ClaimBuster+UKP Mining}: For topic-filtering, we use an argument classification model \cite{reimers2019classification} that has been trained on the UKP Sentential Argument Mining corpus \cite{stab2018cross}. This corpus contains arguments about eight controversial topics annotated over 400 documents. The model takes the topic and sentence as input and gives a score of how related the argument is to the topic. We use this model in a zero-shot setting by feeding in the coronavirus related topics at test-time to compute score for each topic. \\

\paragraph{Step 2. ClaimBuster+Zero-shot NLI:} Following \citet{yin2019benchmarking}, we use pre-trained NLI models as zero-shot text classifiers: we pose the claim sentence to be classified as the NLI premise and construct a hypothesis from each candidate topic. Figure~\ref{fig:mnli_topic_detection} shows the hypothesis corresponding to each of the topics. We then use the entailment score for each topic as its topic score and choose the highest topic score for threshold-based filtering.

\subsection{Claim Object Detection}
\label{sec:claim_object_baseline}

Given the claim sentence and a topic, claim object detection seeks to identify what is being claimed about the topic, as shown in Table~\ref{tab:ex_claim_object}.  %Given the unique nature of this sub-task, we explore it in a few-shot scenario, where we use generative language models to output the claim object. 
We explore this \subtask{} in both zero-shot and few-shot settings by converting it into a prompting task for pre-trained language models as described below:

\paragraph{In-context learning (few-shot):} This setting is similar to \cite{brown2020language}, where the few-shot labeled examples are inserted into the context of a pre-trained language model. The example for which a prediction is to be made is included as a prompt at the end of the context. We refer the reader to Section \ref{sec:app_gpt3} in the appendix for an example. We use GPT-3 \cite{brown2020language} as the language model in this setting.

\paragraph{Prompt-based fine-tuning (few-shot):}  Following \citet{gao2021making}, we fine-tune a pre-trained language model, base-T5 \cite{2020t5}, to learn from a few labeled examples. We convert the examples into a prompt with a format similar to the language model pre-training, which for this model involves generating the target text that has been replaced with a $<$MASK$>$ token in the input. Thus, we convert the few-shot data into such prompts and generate the claim object from the $<$MASK$>$ token. For example, given the claim sentence: \textit{Research conducted on the origin of the virus shows that it came from bats}, and its topic (origin of the virus), the prompt would be: \textit{Research conducted on the origin of the virus shows that it came from bats. The origin of the virus is $<$MASK$>$}.

\paragraph{Prompting (zero-shot):} We consider the language models that were used in few-shot settings above with the same prompts but in zero-shot settings here. In this case, GPT-3 is not provided with any labeled examples in the context and T5 is used out-of-the-box without any fine-tuning. 

\subsection{Stance Detection}
\label{sec:stance_det_baseline}

Given the claim sentence, stance detection identifies if the claimer is asserting or refuting the claim.

\paragraph{Zero-shot NLI:} We leverage NLI models for zero-shot classification. Here, we construct a hypothesis for the \textit{affirm} and the \textit{refute} labels and we take the stance corresponding to a higher entailment score. We consider two settings while constructing the hypothesis based on claim topic availability. Examples are shown in Figure~\ref{fig:mnli_stance_detection}.

\subsection{Claim Span Detection}
\label{sec:claim_span_baseline}
Given a claim sentence, claim span detection identifies the exact claim boundaries within the sentence. 

\paragraph{Debater Boundary Detection:} Our first baseline uses the claim boundary detection service from the Project Debater\footnote{\href{http://research.ibm.com/interactive/project-debater/}{Project Debater}} APIs \cite{bar-haim-etal-2021-project}. This system is based on BERT-Large, which is further fine-tuned on 52K crowd-annotated examples mined from the Lexis-Nexis corpus.\footnote{\href{http://www.lexisnexis.com/en-us/home.page}{http://www.lexisnexis.com/en-us/home.page}} 

\paragraph{PolNeAR-Content:} Our second baseline leverages PolNeAR \cite{newell2018attribution}, a popular news attribution corpus of annotated triples comprising the \textit{source}, a \textit{cue}, and the \textit{content} for statements made in the news. We build a claim span detection model from it by fine-tuning BERT-large \cite{devlin-etal-2019-bert} to identify the \textit{content} span, with a start classifier and an end classifier on top of the encoder outputs, given the sentence as an input.
%to identify the content span, using start and end classifier heads over Heng[I don’t understand what is " start and end classifier heads over"] the encoder outputs,

\subsection{Claimer Detection}
\label{sec:claimer_baseline}

This \subtask{} identifies if the claim is made by the journalist or a reported source, in addition to identifying the mention of the source in the news article.

\paragraph{PolNeAR-Source:} We leverage the PolNeAR corpus to build a claimer extraction baseline. Given a statement, we use the \textit{source} annotation as the claimer and mark the \textit{content} span within the statement using special tokens. We then fine-tune a BERT-large model to extract the source span from the statement using a start classifier and an end classifier over
%\heng{I don't understand what is " start and end classifier heads over"} 
the encoder outputs. At evaluation time, we use the news article as an input, marking the claim span with special tokens and using the sum of the start and the end classifier scores as a claimer span confidence score. This is thresholded to determine if the claim is by the journalist, with the claimer span used as an output for reported claims.

\paragraph{SRL:} We build a Semantic Role Labeling (SRL) baseline for claimer extraction. SRL outputs the verb predicate-argument structure of a sentence such as who did what to whom. Given the claim sentence as an input, we filter out verb predicates that match a pre-defined set of cues\footnote{Appendix \ref{sec:app_srl} contains the complete set of cues.} 
(e.g., \emph{say}, \emph{believe}, \emph{deny}). Then, we use the span corresponding to the ARG-0 (agent) of the predicate as the claimer. As SRL works at the sentence level, this approach cannot extract claimers outside of the claim sentence. Thus, the system outputs \emph{journalist} as the claimer when none of the verb predicates in the sentence matches the pre-defined set of cues. 

%% file: sections/experiments.tex
\section{Experiments}

In this section, we evaluate various zero-shot and few-shot approaches for the \subtask{}s of our claim detection task. To estimate the upper bounds, we also report the human performance for each \subtask{} computed over ten random news articles.

\subsection{Claim Sentence Detection}
\label{sec:exp_claim_detection}

\paragraph{Setup:} 
%For the UKP-Mining topic-filtering system, we use a model based on BERT-large released by \citet{reimers2019classification}  
For zero-shot MNLI, we use BART-large\footnote{\href{http://huggingface.co/facebook/bart-large-mnli}{http://huggingface.co/facebook/bart-large-mnli}} \cite{lewis2020bart} trained on the MultiNLI corpus \cite{williams2018broad}. ClaimBuster and the topic-filtering thresholds are tuned on the development set. For evaluation, we use precision, recall, and F1 scores for the filtered set of claims relative to the ground-truth annotations.

\paragraph{Results and Analysis:} Table~\ref{tab:exp_claim_detection} shows the performance of various systems for identifying claim sentences about COVID-19. We use ClaimBuster, which does not involve topic detection, as a low-precision high-recall baseline. We can see that the performance improves by leveraging a pre-trained NLI model as a zero-shot filter for claims that are not related to the topics at hand. We also report results for both single-human performance and for 3-way majority voting. Note that even humans have relatively lower precision, demonstrating the difficulty of identifying sentences with claims. 
%We hypothesize that this could be due to the subjective nature of whether an assertion is a claim or just a statement.
Nevertheless, the model performance is still considerably worse compared to human performance, showing the need for better models.

\begin{table}[!htb]
\small
    \centering
    \begin{tabular}{l|c|c|c}
      \textbf{Model} & \textbf{P} & \textbf{R} & \textbf{F1}  \\
      \hline
      ClaimBuster & 13.0 & \textbf{86.5} & 22.6 \\
      \hline
      ClaimBuster + Zero-shot NLI & \textbf{21.8} & 53.3 & \textbf{30.9}\\
      \hline
      \hline
      Human (single)& 52.7 & 70.0 & 60.1\\
      \hline
     Human (3-way majority voting)& 60.2 & 83.5 & 70.0\\
      \hline
    \end{tabular}
    \caption{Performance (in \%) for various systems for detecting claims related to COVID-19.}
    %, given a news article as an input.}
    \label{tab:exp_claim_detection}
\end{table}

%It was also evident when we measured the inter-annotator agreement for claim sentence detection, with a Cohen's kappa of 0.44 demonstrating moderate agreement. 

%\heng{This claim performance table is likely to trigger many negative comments. Why is this task so hard, even for human? why human P is so low? For real application the precision needs to be very high. Maybe you can rank the claims based on their frequency and show P/R/F for the top ranked ones? and emphasize this is for zero-shot setting}

% \begin{table}[!htb]
% \small
%     \centering
%     \begin{tabular}{l|c|c|c}
%       \textbf{Model} & \textbf{P} & \textbf{R} & \textbf{F1}  \\
%       \hline
%       ClaimBuster & 6.3 & 71.8 & 11.7 \\
%       \hline
%       ClaimBuster + UKP Mining & 11.3 & 44.5 & 18.1 \\
%       \hline
%       ClaimBuster + Zero-shot MNLI & 14.1 & 42.2 & 21.1\\
%       \hline
%       \hline
%       Human & & & \\
%       \hline
%     \end{tabular}
%     \caption{Performance (in \%) of different systems for detecting claims related to the coronavirus, given the news article as input.}
%     \label{tab:exp_claim_detection}
% \end{table}

%\heng{I think your observation about in-sentence vs. outside sentence is very interesting, maybe emphasize that in analysis}

\subsection{Claim Object Detection}

\paragraph{Setup:} We use the development set to get the few-shot examples, sampling\footnote{We will release the few-shot examples for reproducibility.} five examples per topic. To account for sampling variance, we report numbers averaged over three runs. For language model sizes to be comparable, we use the Ada\footnote{\href{https://blog.eleuther.ai/gpt3-model-sizes/}{https://blog.eleuther.ai/gpt3-model-sizes/}} version of GPT-3 and the base version of T5. We fine-tune T5-base for five epochs using a learning rate of 3e-5. We score using string-match F1, as done for question answering \cite{rajpurkar2016squad}.

\paragraph{Results and Analysis:} Table~\ref{tab:exp_object_extraction} shows the F1 score for extracting the claim object related to the topic. In zero-shot settings, we see that GPT-3 performs considerably better than T5, potentially benefiting from the larger corpus it was trained on. However, in a few-shot setting, T5 is competitive with GPT-3, showing the promise of prompt-based fine-tuning, even with limited few-shot examples.
%We can see that using such prompt-based fine-tuning approaches with smaller models (220M) can considerably outperform in-context learning approach using much larger models (1.3B), demonstrating the effectiveness of pattern-based fine-tuning.

%\begin{table*}[!htb]
%\small
%    \centering
%    \begin{tabular}{c|c|c|c|c|c}
%     \textbf{Model} & \textbf{Origin} & \textbf{Transmission} & \textbf{Prevention} & \textbf{Cure} &  \textbf{Overall}\\
%     \hline
%     Prompt-based fine-tuning (T5-base) & 31.1 & 30.1 & 61.1 & \textbf{70.1} & 38.8 \\
%     \hline
%     Prompt-based fine-tuning (T5-large) &  &  &  &  &  \\
%     \hline
%     In-context learning (GPT-neo) & \textbf{33.8} & \textbf{75.0} & \textbf{69.8} & 50.8 & \textbf{44.2} \\
%     \hline
%     In-context learning (GPT-3) & &  &  &  &  \\
%     \hline
%     \hline
%     Human & & & & & \\
%     \hline
%    \end{tabular}
%    \caption{Topic-wise F1 (in \%) and overall F1 (in \%) of different few-shot systems for the claim object extraction sub-task.}
%    \label{tab:exp_object_extraction}
%\end{table*}

\begin{table}[!htb]
\small
    \centering
    \begin{tabular}{l|c|c|c}
     \textbf{Approach} & \textbf{Model} & \textbf{Type} & \textbf{F1}\\
     \hline
     Prompting & GPT-3 & Zero-shot & 15.2\\
     \hline
     Prompting & T5 & Zero-shot & 11.4 \\
     \hline
     In-context learning & GPT-3 & Few-Shot & \textbf{51.9} \\
     \hline
     Prompt-based fine-tuning & T5 & Few-Shot & 51.6 \\
     \hline
     \hline
     Human & - & - & 67.7 \\
     \hline
    \end{tabular}
    \caption{F1 score (in \%) for various zero-shot and few-shot systems for the claim object detection sub-task.}
    \label{tab:exp_object_extraction}
\end{table}

\subsection{Stance Detection}

\paragraph{Setup:}  We use the same BART-large model trained for NLI as in Section~\ref{sec:exp_claim_detection}. In the setting with access to the topic, we take the topic from the gold-standard annotation.

\paragraph{Results and Analysis:} We also consider a majority class baseline that always predicts \emph{affirm} as the stance. Table~\ref{tab:exp_stance_detection} shows the performance of stance detection approaches. We can see that the the NLI model with access to the topic performs the best, with considerable improvement in performance for the \emph{refute} class. Thus, access to additional attribute information helps here as the topic of the claim can be used to come up with a more relevant hypothesis, as is evident from Figure~\ref{fig:mnli_stance_detection}.

\begin{table}[!htb]
\small
    \centering
    \begin{tabular}{l|c|c|c}
    \textbf{Model} & \textbf{Affirm F1} & \textbf{Refute F1} & \textbf{Acc.} \\
    \hline
    Majority class &  82.5 & 0.0 & 70.3 \\
    \hline
    NLI (no topic) & 89.1 & 68.0 & 83.8 \\
    \hline
    NLI (with topic) & \textbf{91.1} & \textbf{78.8} & \textbf{87.5} \\
    \hline
    \hline
    Human & 97.0 & 84.2 & 94.9 \\
    \hline
    \end{tabular}
    \caption{F1 score (in \%) for \emph{affirm} and \emph{refute} classes along with the overall accuracy for stance detection. Zero-shot NLI is shown separately based on access the topic while constructing the hypothesis.}
    \label{tab:exp_stance_detection}
\end{table}

\subsection{Claim Span Detection}

\paragraph{Results and Analysis:}  The evaluation measure in this setting is character-span F1. From Table~\ref{tab:exp_span_detection}, we see that the Debater claim boundary detection system considerably outperforms the attribution-based system. This could be because the former is trained on arguments, which are more similar to claims compared to statement-like attributions.

\begin{table}[!htb]
\small
    \centering
    \begin{tabular}{l|c|c|c}
      \textbf{Model} & \textbf{Prec.} & \textbf{Recall} & \textbf{F1}  \\
      \hline
      PolNeAR-Content & 67.0 & 42.8 & 52.3 \\
      \hline
      Debater Boundary Detection & \textbf{75.7} & \textbf{77.7} & \textbf{76.7} \\
      \hline
      \hline
      Human & 82.7 & 90.9 & 86.6 \\
      \hline
    \end{tabular}
    \caption{Performance (in \%) of different systems for identifying the boundaries of the claim.} 
%    within a given claim sentence.}
    \label{tab:exp_span_detection}
\end{table}

\subsection{Claimer Detection}

\paragraph{Setup:} For the PolNeAR-Source system, the threshold for confidence score is tuned on the dev set. The claim span output from the Debater boundary system is used for marking the claim content in the context. For the SRL system, we leverage the parser\footnote{\href{http://docs.allennlp.org/models/main/models/structured\_prediction/predictors/srl/}{AllenNLP SRL Parser}} provided by AllenNLP \cite{gardner2018allennlp}, which was trained on OntoNotes \cite{pradhan-etal-2013-towards}. The evaluation involves scores for the journalist (classification F1) and for reported (string-match F1), along with overall F1.

\begin{table}[!htb]
\small
    \centering
    \begin{tabular}{l|c|c|c}
      \textbf{Model} &\textbf{F1} & \textbf{Reported} & \textbf{Journalist}   \\
      \hline
      SRL & 41.7 & 23.5  & \textbf{67.2} \\
      \hline
      PolNeAR-Source & \textbf{42.3} & \textbf{25.5} & 65.9 \\
      \hline
      \hline
      Human & 85.8 & 81.3 & 88.9\\
      \hline
    \end{tabular}
    \caption{Claimer detection scores for journalist claims and for reported claims, along with the overall F1.}
    \label{tab:exp_claimer_extraction}
\end{table}

\paragraph{Results and Analysis:}
Table \ref{tab:exp_claimer_extraction} shows that automatic models perform considerably worse than humans for claimer detection. While the performance is relatively better for identifying whether a journalist is making the claim, models perform poorly for reported claims, which involves extracting the claimer mentions. For reported claims, Table~\ref{tab:exp_claimer_extraction_by_loc} shows that the performance depends on whether the claimer is mentioned inside or outside of the claim sentence. Specifically, we see that these attribution models are able to handle claimer detection for reported claims only when the claimer mention is within the claim sentence.
The need for cross-sentence reasoning for the claimer detection sub-task is evident from the low out-of-sentence F1 score for these sentence-level approaches. 

\begin{table}[!htb]
\small
    \centering
    \begin{tabular}{l|c|c}
      \textbf{Model} & \textbf{In-sentence} & \textbf{Out-of-sentence}   \\
      \hline
      SRL & 35.8 & 2.4  \\
      \hline
      PolNeAR-Source & \textbf{38.9} & \textbf{2.7} \\
      \hline
    \end{tabular}
    \caption{F1 score (in \%) in terms of reported claims for extracting the claimer when it is present within 
%    (in-sentence) 
    or outside
%    (out-of-sentence) 
    the claim sentence.}
    \label{tab:exp_claimer_extraction_by_loc}
\end{table}

\subsection{Error Analysis and Remaining Challenges}
\begin{figure}[]
    \centering
    \begin{subfigure}[c]{1.0\linewidth}
        \centering
         \includegraphics[width=1\linewidth]{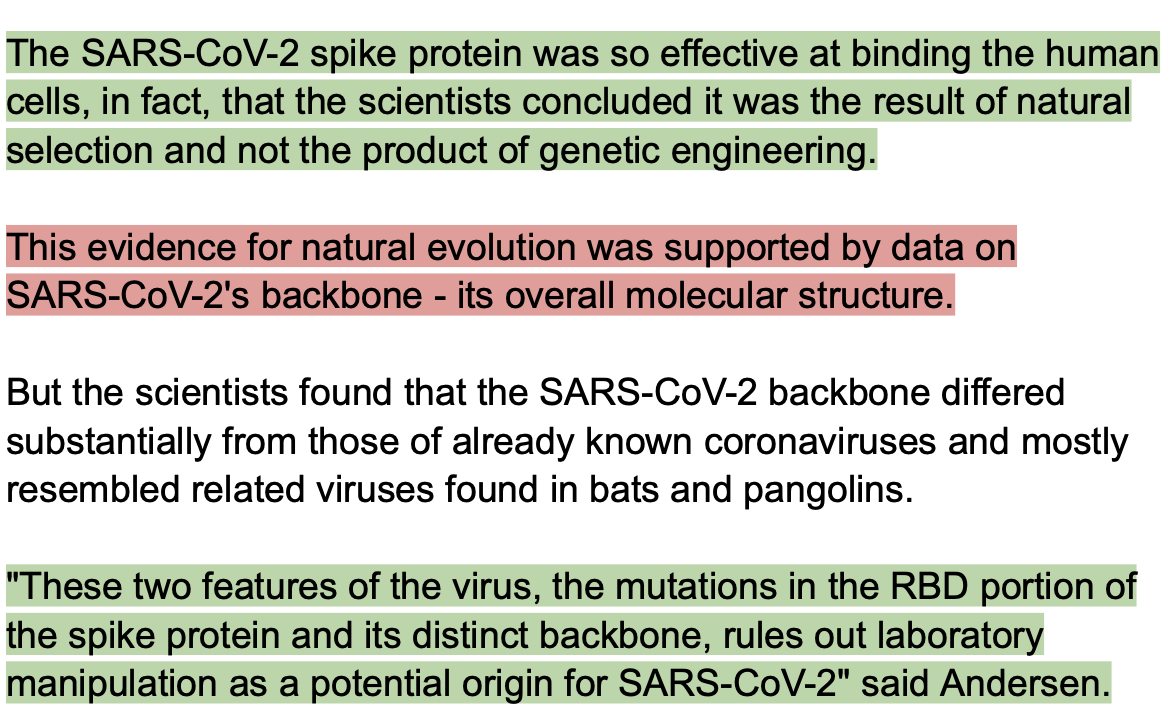}
           \vspace{-1.5em}
           \caption{}
           \label{fig:human_error_1}
    \end{subfigure}
    \hfill
    \begin{subfigure}[c]{1.0\linewidth}
        \centering
         \includegraphics[width=1\linewidth]{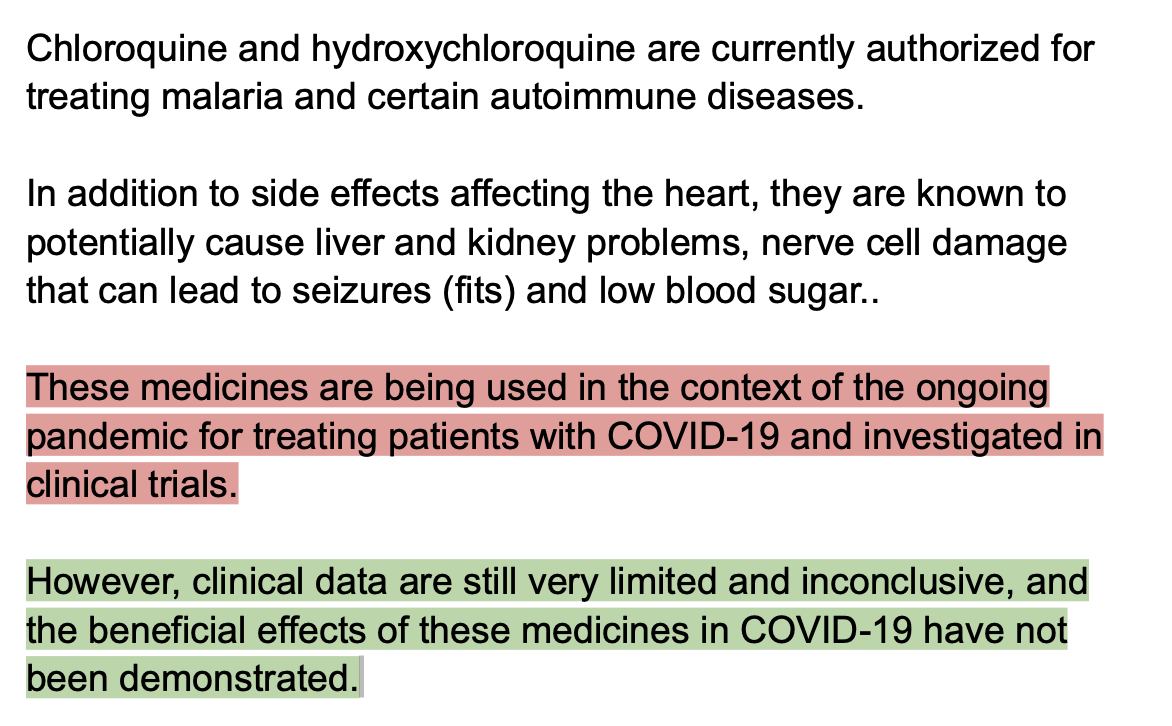}
           \vspace{-1.5em}
           \caption{}
           \label{fig:human_error_2}
    \end{subfigure}
    \vspace{-0.5em}
    \caption{Some examples from the human study with the gold-standard claims highlighted in green and false positives from humans highlighted in red.}
    \label{fig:human_errors}
\end{figure}

News articles have a narrative structure when presenting claims, by backing them up with some evidence. We observed that humans, when considering sentences without looking at the context, tend to identify such statements providing evidential information as claims too. Figure \ref{fig:human_errors} shows some examples of errors corresponding to false positives from the human study. The human study identified the sentences in red as claims, in addition to the ones in green. In Figure \ref{fig:human_error_1}, the sentences in green contain concrete claims regarding the origin of the virus, with the first sentence claiming that it came from natural selection and the second sentence refuting that the virus was a laboratory manipulation. The sentence in red, on the other hand, simply provides evidence for natural evolution. In Figure \ref{fig:human_error_2}, the sentence in green contains a claim that refutes that these medicines can cure the virus. On the other hand, the sentence in red does not contain a claim because it simply asserts that these medicines are being used for treating patients, without any clear claim on whether they can actually cure the virus.

\begin{table*}[]
    \centering
    \small
    \begin{tabular}{p{35em}|c|c}
   \centering \textbf{Claim Sentence}  &  \textbf{Gold topic} & \textbf{Predicted topic} \\
    \hline
    This novel coronavirus was believed to have started in a large seafood or wet market, suggesting animal-to-person spread. & \multirow{2}{*}{Origin} &  \multirow{2}{*}{Transmission}\\
     \hline
    A Wuhan laboratory official has denied any role in spreading the new coronavirus, after months of speculation about how the previously unknown animal disease made the leap to humans. &  \multirow{3}{*}{Origin}  & \multirow{3}{*}{Transmission} \\
     \hline
    One medication, an antiviral drug called Remdesivir, has been shown in certain studies to improve symptoms and shorten hospital stays. & \multirow{2}{*}{Cure}  & \multirow{2}{*}{Protection}\\
     \hline
    Studies show hydroxychloroquine does not have clinical benefits in treating COVID-19. & Cure   & Protection \\
     \hline
    \end{tabular}
    \caption{Some topic classification error examples from the zero-shot NLI model.}
    \label{tab:topic_errors}
\end{table*}

\begin{figure}[]
    \centering
    \includegraphics[width=0.9\linewidth]{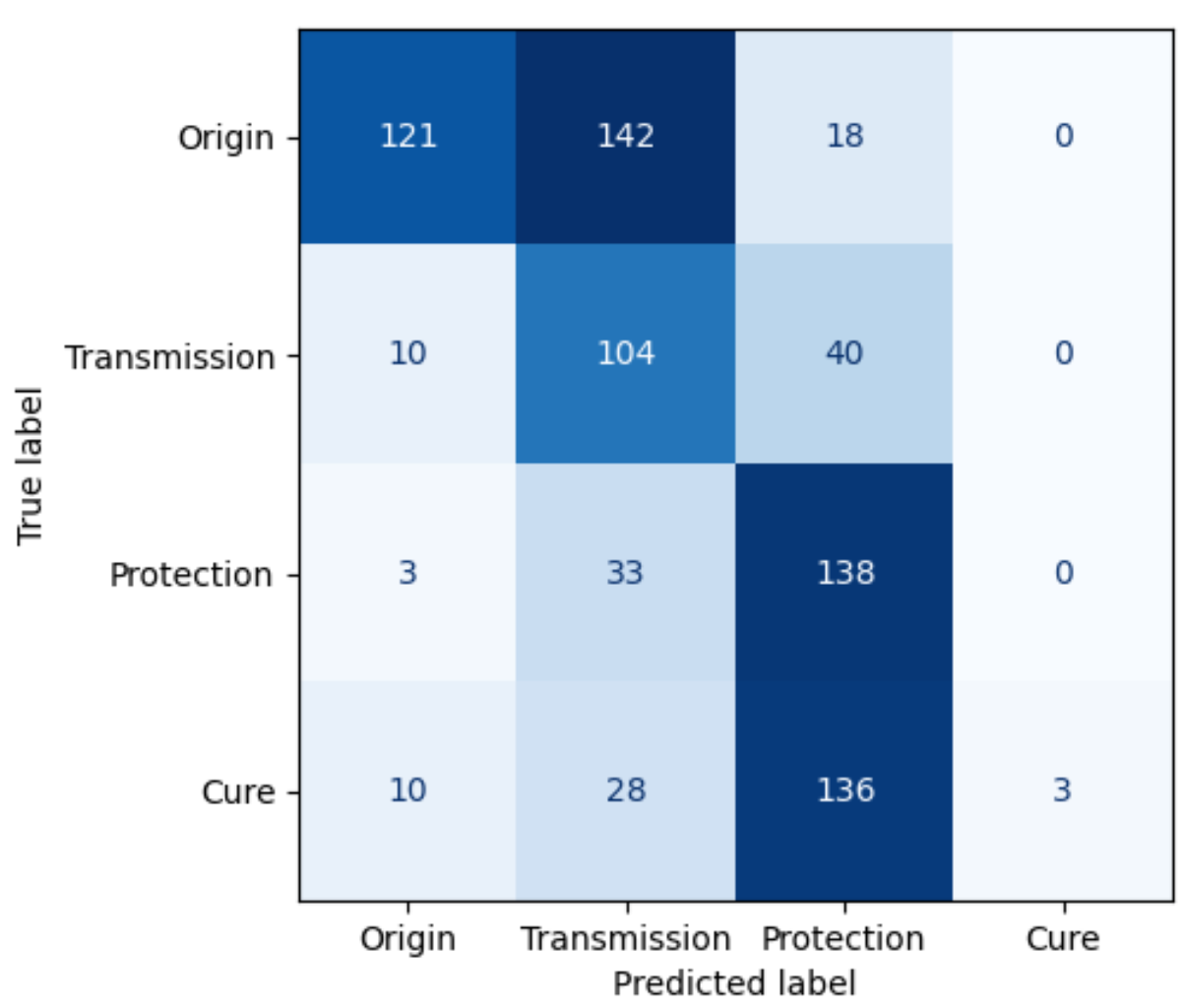}
    \caption{Confusion matrix for the topic classification predictions from the zero-shot NLI model.}
    \label{fig:topic_confusion}
\end{figure}

We investigated the NLI model performance for topic classification. Given the gold-standard claim sentence, the accuracy is 46.6\% over these four topics. Topic-wise F1 was relatively poor for \emph{Cure} (3.3\%) compared to the other topics: \emph{Origin} is 56.9\%, \emph{Protection} is 54.5\%, and \emph{Transmission} is 45.1\%. Figure \ref{fig:topic_confusion} shows the confusion matrix for the topic classification predictions. We see two dominant types of errors. First, most claims corresponding to the topic \emph{Cure} are under \emph{Protection}. This is potentially due to these two topics being related and the NLI model unable to differentiate that \emph{Protection} corresponds to prevention measures before contracting COVID-19, while \emph{Cure} refers to treatments after contracting COVID-19. Second, we see that a considerable number of claims related to \emph{Origin} were classified as \emph{Transmission}. This could be due to a statement about the virus originating in animals and then jumping to humans, which suggests that a claim about the origin of the virus was being misconstrued as one regarding the transmission of the virus. Some representative examples for both of these types of errors are shown in Table~\ref{tab:topic_errors}. Given the low topic classification performance of the NLI model, we need better zero-shot approaches for selecting claims related to COVID-19. This is important as the claim topic is crucial to claim object detection and it can help stance detection.

Stance detection performance could likely be improved by also leveraging claim objects while formulating the NLI hypothesis. For example, the stance for \textit{``An Oxford University professor claimed that the coronavirus may not have originated in China.''} was predicted as \emph{affirm} even though it refutes that the virus originated in China. By leveraging the extracted claim object, the NLI hypothesis for the \emph{refute} class could be better formulated as \textit{``China is not the origin of the virus''}. The existing formulation, shown in Figure~\ref{fig:mnli_stance_detection}, only uses the claim topic to put it as \textit{``This refutes the origin of the virus''}. We leave this for future work.

The claimer detection \subtask{} requires incorporating stronger cross-sentence reasoning
%to identify the claimer for 
when the mention is outside the claim sentence. This requires building attribution systems that are document-level. Moreover, the same news article can have similar claims but from different claimers. To prevent misattribution in such cases, it would be beneficial to identify the context within the news article that is relevant to the given claim, so as to remove noise from other related claims.

%% file: emnlp2022.bbl
\begin{thebibliography}{49}
\expandafter\ifx\csname natexlab\endcsname\relax\def\natexlab#1{#1}\fi

\bibitem[{Alam et~al.(2020)Alam, Shaar, Dalvi, Sajjad, Nikolov, Mubarak,
  Martino, Abdelali, Durrani, Darwish et~al.}]{alam2020fighting}
Firoj Alam, Shaden Shaar, Fahim Dalvi, Hassan Sajjad, Alex Nikolov, Hamdy
  Mubarak, Giovanni Da~San Martino, Ahmed Abdelali, Nadir Durrani, Kareem
  Darwish, et~al. 2020.
\newblock Fighting the covid-19 infodemic: modeling the perspective of
  journalists, fact-checkers, social media platforms, policy makers, and the
  society.
\newblock \emph{arXiv preprint arXiv:2005.00033}.

\bibitem[{Allaway and McKeown(2020)}]{allaway-mckeown-2020-zero}
Emily Allaway and Kathleen McKeown. 2020.
\newblock \href {https://doi.org/10.18653/v1/2020.emnlp-main.717}
  {{Z}ero-{S}hot {S}tance {D}etection: {A} {D}ataset and {M}odel using
  {G}eneralized {T}opic {R}epresentations}.
\newblock In \emph{Proceedings of the 2020 Conference on Empirical Methods in
  Natural Language Processing (EMNLP)}, pages 8913--8931, Online. Association
  for Computational Linguistics.

\bibitem[{Aly et~al.(2021)Aly, Guo, Schlichtkrull, Thorne, Vlachos,
  Christodoulopoulos, Cocarascu, and Mittal}]{aly2021feverous}
Rami Aly, Zhijiang Guo, Michael Schlichtkrull, James Thorne, Andreas Vlachos,
  Christos Christodoulopoulos, Oana Cocarascu, and Arpit Mittal. 2021.
\newblock Feverous: Fact extraction and verification over unstructured and
  structured information.
\newblock \emph{arXiv preprint arXiv:2106.05707}.

\bibitem[{Arslan et~al.(2020)Arslan, Hassan, Li, and
  Tremayne}]{arslan2020benchmark}
Fatma Arslan, Naeemul Hassan, Chengkai Li, and Mark Tremayne. 2020.
\newblock A benchmark dataset of check-worthy factual claims.
\newblock In \emph{Proceedings of the International AAAI Conference on Web and
  Social Media}, volume~14, pages 821--829.

\bibitem[{Atanasova et~al.(2019{\natexlab{a}})Atanasova, Nakov, Karadzhov,
  Mohtarami, and Da~San~Martino}]{atanasova2019overview}
Pepa Atanasova, Preslav Nakov, Georgi Karadzhov, Mitra Mohtarami, and Giovanni
  Da~San~Martino. 2019{\natexlab{a}}.
\newblock Overview of the clef-2019 checkthat! lab: Automatic identification
  and verification of claims. task 1: Check-worthiness.
\newblock \emph{CLEF (Working Notes)}, 2380.

\bibitem[{Atanasova et~al.(2019{\natexlab{b}})Atanasova, Nakov, M{\`a}rquez,
  Barr{\'o}n-Cede{\~n}o, Karadzhov, Mihaylova, Mohtarami, and
  Glass}]{atanasova2019automatic}
Pepa Atanasova, Preslav Nakov, Llu{\'\i}s M{\`a}rquez, Alberto
  Barr{\'o}n-Cede{\~n}o, Georgi Karadzhov, Tsvetomila Mihaylova, Mitra
  Mohtarami, and James Glass. 2019{\natexlab{b}}.
\newblock Automatic fact-checking using context and discourse information.
\newblock \emph{Journal of Data and Information Quality (JDIQ)}, 11(3):1--27.

\bibitem[{Augenstein et~al.(2019)Augenstein, Lioma, Wang, Lima, Hansen, Hansen,
  and Simonsen}]{augenstein2019multifc}
Isabelle Augenstein, Christina Lioma, Dongsheng Wang, Lucas~Chaves Lima, Casper
  Hansen, Christian Hansen, and Jakob~Grue Simonsen. 2019.
\newblock Multifc: A real-world multi-domain dataset for evidence-based fact
  checking of claims.
\newblock In \emph{Proceedings of the 2019 Conference on Empirical Methods in
  Natural Language Processing and the 9th International Joint Conference on
  Natural Language Processing (EMNLP-IJCNLP)}, pages 4685--4697.

\bibitem[{Bar-Haim et~al.(2020)Bar-Haim, Eden, Friedman, Kantor, Lahav, and
  Slonim}]{bar2020arguments}
Roy Bar-Haim, Lilach Eden, Roni Friedman, Yoav Kantor, Dan Lahav, and Noam
  Slonim. 2020.
\newblock From arguments to key points: Towards automatic argument
  summarization.
\newblock In \emph{Proceedings of the 58th Annual Meeting of the Association
  for Computational Linguistics}, pages 4029--4039.

\bibitem[{Bar-Haim et~al.(2021)Bar-Haim, Kantor, Venezian, Katz, and
  Slonim}]{bar-haim-etal-2021-project}
Roy Bar-Haim, Yoav Kantor, Elad Venezian, Yoav Katz, and Noam Slonim. 2021.
\newblock \href {https://aclanthology.org/2021.emnlp-demo.31} {Project
  {D}ebater {API}s: {D}ecomposing the {AI} grand challenge}.
\newblock In \emph{Proceedings of the 2021 Conference on Empirical Methods in
  Natural Language Processing: System Demonstrations}, pages 267--274, Online
  and Punta Cana, Dominican Republic. Association for Computational
  Linguistics.

\bibitem[{Brown et~al.(2020)Brown, Mann, Ryder, Subbiah, Kaplan, Dhariwal,
  Neelakantan, Shyam, Sastry, Askell et~al.}]{brown2020language}
Tom~B Brown, Benjamin Mann, Nick Ryder, Melanie Subbiah, Jared Kaplan, Prafulla
  Dhariwal, Arvind Neelakantan, Pranav Shyam, Girish Sastry, Amanda Askell,
  et~al. 2020.
\newblock Language models are few-shot learners.
\newblock \emph{arXiv preprint arXiv:2005.14165}.

\bibitem[{Buhrmester et~al.(2011)Buhrmester, Kwang, and
  Gosling}]{buhrmester2011amazon}
Michael Buhrmester, Tracy Kwang, and Samuel~D Gosling. 2011.
\newblock Amazon's mechanical turk: A new source of inexpensive, yet
  high-quality, data?
\newblock \emph{Perspectives on Psychological Science}, pages 3--5.

\bibitem[{Devlin et~al.(2019)Devlin, Chang, Lee, and
  Toutanova}]{devlin-etal-2019-bert}
Jacob Devlin, Ming-Wei Chang, Kenton Lee, and Kristina Toutanova. 2019.
\newblock \href {https://doi.org/10.18653/v1/N19-1423} {{BERT}: Pre-training of
  deep bidirectional transformers for language understanding}.
\newblock In \emph{Proceedings of the 2019 Conference of the North {A}merican
  Chapter of the Association for Computational Linguistics: Human Language
  Technologies, Volume 1 (Long and Short Papers)}, pages 4171--4186,
  Minneapolis, Minnesota. Association for Computational Linguistics.

\bibitem[{Elson and McKeown(2010)}]{elson2010automatic}
David~K Elson and Kathleen~R McKeown. 2010.
\newblock Automatic attribution of quoted speech in literary narrative.
\newblock In \emph{Twenty-Fourth AAAI Conference on Artificial Intelligence}.

\bibitem[{Gao et~al.(2021)Gao, Fisch, and Chen}]{gao2021making}
Tianyu Gao, Adam Fisch, and Danqi Chen. 2021.
\newblock Making pre-trained language models better few-shot learners.
\newblock In \emph{Association for Computational Linguistics (ACL)}.

\bibitem[{Gardner et~al.(2018)Gardner, Grus, Neumann, Tafjord, Dasigi, Liu,
  Peters, Schmitz, and Zettlemoyer}]{gardner2018allennlp}
Matt Gardner, Joel Grus, Mark Neumann, Oyvind Tafjord, Pradeep Dasigi, Nelson~F
  Liu, Matthew Peters, Michael Schmitz, and Luke Zettlemoyer. 2018.
\newblock Allennlp: A deep semantic natural language processing platform.
\newblock In \emph{Proceedings of Workshop for NLP Open Source Software
  (NLP-OSS)}, pages 1--6.

\bibitem[{Gencheva et~al.(2017)Gencheva, Nakov, M{\`a}rquez,
  Barr{\'o}n-Cede{\~n}o, and Koychev}]{gencheva-etal-2017-context}
Pepa Gencheva, Preslav Nakov, Llu{\'\i}s M{\`a}rquez, Alberto
  Barr{\'o}n-Cede{\~n}o, and Ivan Koychev. 2017.
\newblock \href {https://doi.org/10.26615/978-954-452-049-6_037} {A
  context-aware approach for detecting worth-checking claims in political
  debates}.
\newblock In \emph{Proceedings of the International Conference Recent Advances
  in Natural Language Processing, {RANLP} 2017}, Varna, Bulgaria. INCOMA Ltd.

\bibitem[{Hanselowski et~al.(2019)Hanselowski, Stab, Schulz, Li, and
  Gurevych}]{hanselowski2019richly}
Andreas Hanselowski, Christian Stab, Claudia Schulz, Zile Li, and Iryna
  Gurevych. 2019.
\newblock A richly annotated corpus for different tasks in automated
  fact-checking.
\newblock In \emph{Proceedings of the 23rd Conference on Computational Natural
  Language Learning (CoNLL)}, pages 493--503.

\bibitem[{Hardalov et~al.(2021{\natexlab{a}})Hardalov, Arora, Nakov, and
  Augenstein}]{hardalov-etal-2021-cross}
Momchil Hardalov, Arnav Arora, Preslav Nakov, and Isabelle Augenstein.
  2021{\natexlab{a}}.
\newblock \href {https://aclanthology.org/2021.emnlp-main.710} {Cross-domain
  label-adaptive stance detection}.
\newblock In \emph{Proceedings of the 2021 Conference on Empirical Methods in
  Natural Language Processing}, pages 9011--9028, Online and Punta Cana,
  Dominican Republic. Association for Computational Linguistics.

\bibitem[{Hardalov et~al.(2021{\natexlab{b}})Hardalov, Arora, Nakov, and
  Augenstein}]{hardalov2021survey}
Momchil Hardalov, Arnav Arora, Preslav Nakov, and Isabelle Augenstein.
  2021{\natexlab{b}}.
\newblock A survey on stance detection for mis-and disinformation
  identification.
\newblock \emph{arXiv preprint arXiv:2103.00242}.

\bibitem[{Hassan et~al.(2017{\natexlab{a}})Hassan, Arslan, Li, and
  Tremayne}]{hassan2017toward}
Naeemul Hassan, Fatma Arslan, Chengkai Li, and Mark Tremayne.
  2017{\natexlab{a}}.
\newblock Toward automated fact-checking: Detecting check-worthy factual claims
  by claimbuster.
\newblock In \emph{Proceedings of the 23rd ACM SIGKDD International Conference
  on Knowledge Discovery and Data Mining}, pages 1803--1812.

\bibitem[{Hassan et~al.(2017{\natexlab{b}})Hassan, Zhang, Arslan, Caraballo,
  Jimenez, Gawsane, Hasan, Joseph, Kulkarni, Nayak
  et~al.}]{hassan2017claimbuster}
Naeemul Hassan, Gensheng Zhang, Fatma Arslan, Josue Caraballo, Damian Jimenez,
  Siddhant Gawsane, Shohedul Hasan, Minumol Joseph, Aaditya Kulkarni,
  Anil~Kumar Nayak, et~al. 2017{\natexlab{b}}.
\newblock Claimbuster: The first-ever end-to-end fact-checking system.
\newblock \emph{Proceedings of the VLDB Endowment}, 10(12):1945--1948.

\bibitem[{Jaradat et~al.(2018)Jaradat, Gencheva, Barr{\'o}n-Cede{\~n}o,
  M{\`a}rquez, and Nakov}]{jaradat-etal-2018-claimrank}
Israa Jaradat, Pepa Gencheva, Alberto Barr{\'o}n-Cede{\~n}o, Llu{\'\i}s
  M{\`a}rquez, and Preslav Nakov. 2018.
\newblock \href {https://aclanthology.org/N18-5006} {{C}laim{R}ank: Detecting
  check-worthy claims in {A}rabic and {E}nglish}.
\newblock In \emph{Proceedings of the 2018 Conference of the North {A}merican
  Chapter of the Association for Computational Linguistics: Demonstrations},
  New Orleans, Louisiana. Association for Computational Linguistics.

\bibitem[{Jiang et~al.(2021)Jiang, Song, Scarton, Aker, and
  Bontcheva}]{jiang2021categorising}
Ye~Jiang, Xingyi Song, Carolina Scarton, Ahmet Aker, and Kalina Bontcheva.
  2021.
\newblock Categorising fine-to-coarse grained misinformation: An empirical
  study of covid-19 infodemic.
\newblock \emph{arXiv preprint arXiv:2106.11702}.

\bibitem[{Karadzhov et~al.(2017)Karadzhov, Nakov, M{\`a}rquez,
  Barr{\'o}n-Cede{\~n}o, and Koychev}]{karadzhov2017fully}
Georgi Karadzhov, Preslav Nakov, Llu{\'\i}s M{\`a}rquez, Alberto
  Barr{\'o}n-Cede{\~n}o, and Ivan Koychev. 2017.
\newblock Fully automated fact checking using external sources.
\newblock \emph{arXiv preprint arXiv:1710.00341}.

\bibitem[{Kotonya and Toni(2020)}]{kotonya2020explainable}
Neema Kotonya and Francesca Toni. 2020.
\newblock Explainable automated fact-checking for public health claims.
\newblock In \emph{Proceedings of the 2020 Conference on Empirical Methods in
  Natural Language Processing (EMNLP)}, pages 7740--7754.

\bibitem[{Levy et~al.(2014)Levy, Bilu, Hershcovich, Aharoni, and
  Slonim}]{levy2014context}
Ran Levy, Yonatan Bilu, Daniel Hershcovich, Ehud Aharoni, and Noam Slonim.
  2014.
\newblock Context dependent claim detection.
\newblock In \emph{Proceedings of COLING 2014, the 25th International
  Conference on Computational Linguistics: Technical Papers}, pages 1489--1500.

\bibitem[{Levy et~al.(2017)Levy, Gretz, Sznajder, Hummel, Aharonov, and
  Slonim}]{levy2017unsupervised}
Ran Levy, Shai Gretz, Benjamin Sznajder, Shay Hummel, Ranit Aharonov, and Noam
  Slonim. 2017.
\newblock Unsupervised corpus--wide claim detection.
\newblock In \emph{Proceedings of the 4th Workshop on Argument Mining}, pages
  79--84.

\bibitem[{Lewis et~al.(2020)Lewis, Liu, Goyal, Ghazvininejad, Mohamed, Levy,
  Stoyanov, and Zettlemoyer}]{lewis2020bart}
Mike Lewis, Yinhan Liu, Naman Goyal, Marjan Ghazvininejad, Abdelrahman Mohamed,
  Omer Levy, Veselin Stoyanov, and Luke Zettlemoyer. 2020.
\newblock Bart: Denoising sequence-to-sequence pre-training for natural
  language generation, translation, and comprehension.
\newblock In \emph{Proceedings of the 58th Annual Meeting of the Association
  for Computational Linguistics}, pages 7871--7880.

\bibitem[{Liu et~al.(2021)Liu, Yuan, Fu, Jiang, Hayashi, and
  Neubig}]{liu2021pre}
Pengfei Liu, Weizhe Yuan, Jinlan Fu, Zhengbao Jiang, Hiroaki Hayashi, and
  Graham Neubig. 2021.
\newblock Pre-train, prompt, and predict: A systematic survey of prompting
  methods in natural language processing.
\newblock \emph{arXiv preprint arXiv:2107.13586}.

\bibitem[{Naeem and Bhatti(2020)}]{naeem2020covid}
Salman~Bin Naeem and Rubina Bhatti. 2020.
\newblock The covid-19 ‘infodemic’: a new front for information
  professionals.
\newblock \emph{Health Information \& Libraries Journal}, 37(3):233--239.

\bibitem[{Nakov et~al.(2022)Nakov, Barr{\'{o}}n{-}Cede{\~{n}}o, Martino, Alam,
  Stru{\ss}, Mandl, M{\'{\i}}guez, Caselli, Kutlu, Zaghouani, Li, Shaar, Shahi,
  Mubarak, Nikolov, Babulkov, Kartal, and
  Beltr{\'{a}}n}]{CLEF2022:CheckThat:ECIR}
Preslav Nakov, Alberto Barr{\'{o}}n{-}Cede{\~{n}}o, Giovanni Da~San Martino,
  Firoj Alam, Julia~Maria Stru{\ss}, Thomas Mandl, Rub{\'{e}}n M{\'{\i}}guez,
  Tommaso Caselli, M{\"{u}}cahid Kutlu, Wajdi Zaghouani, Chengkai Li, Shaden
  Shaar, Gautam~Kishore Shahi, Hamdy Mubarak, Alex Nikolov, Nikolay Babulkov,
  Yavuz~Selim Kartal, and Javier Beltr{\'{a}}n. 2022.
\newblock \href {https://doi.org/10.1007/978-3-030-99739-7\_52} {The
  {CLEF-2022} checkthat! lab on fighting the {COVID-19} infodemic and fake news
  detection}.
\newblock In \emph{Advances in Information Retrieval - 44th European Conference
  on {IR} Research, {ECIR} 2022, Stavanger, Norway, April 10-14, 2022,
  Proceedings, Part {II}}, volume 13186 of \emph{Lecture Notes in Computer
  Science}, pages 416--428. Springer.

\bibitem[{Nakov et~al.(2021)Nakov, Corney, Hasanain, Alam, Elsayed,
  Barr{\'o}n-Cede{\~n}o, Papotti, Shaar, and Martino}]{nakov2021automated}
Preslav Nakov, David Corney, Maram Hasanain, Firoj Alam, Tamer Elsayed, Alberto
  Barr{\'o}n-Cede{\~n}o, Paolo Papotti, Shaden Shaar, and Giovanni Da~San
  Martino. 2021.
\newblock Automated fact-checking for assisting human fact-checkers.
\newblock \emph{arXiv preprint arXiv:2103.07769}.

\bibitem[{Newell et~al.(2018)Newell, Margolin, and
  Ruths}]{newell2018attribution}
Edward Newell, Drew Margolin, and Derek Ruths. 2018.
\newblock An attribution relations corpus for political news.
\newblock In \emph{Proceedings of the Eleventh International Conference on
  Language Resources and Evaluation (LREC 2018)}.

\bibitem[{Palau and Moens(2009)}]{palau2009argumentation}
Raquel~Mochales Palau and Marie-Francine Moens. 2009.
\newblock Argumentation mining: the detection, classification and structure of
  arguments in text.
\newblock In \emph{Proceedings of the 12th international conference on
  artificial intelligence and law}, pages 98--107.

\bibitem[{Pareti(2016{\natexlab{a}})}]{pareti2016parc}
Silvia Pareti. 2016{\natexlab{a}}.
\newblock Parc 3.0: A corpus of attribution relations.
\newblock In \emph{Proceedings of the Tenth International Conference on
  Language Resources and Evaluation (LREC'16)}, pages 3914--3920.

\bibitem[{Pareti(2016{\natexlab{b}})}]{pareti-2016-parc}
Silvia Pareti. 2016{\natexlab{b}}.
\newblock \href {https://aclanthology.org/L16-1619} {{PARC} 3.0: A corpus of
  attribution relations}.
\newblock In \emph{Proceedings of the Tenth International Conference on
  Language Resources and Evaluation ({LREC}'16)}, pages 3914--3920,
  Portoro{\v{z}}, Slovenia. European Language Resources Association (ELRA).

\bibitem[{Pradhan et~al.(2013)Pradhan, Moschitti, Xue, Ng, Bj{\"o}rkelund,
  Uryupina, Zhang, and Zhong}]{pradhan-etal-2013-towards}
Sameer Pradhan, Alessandro Moschitti, Nianwen Xue, Hwee~Tou Ng, Anders
  Bj{\"o}rkelund, Olga Uryupina, Yuchen Zhang, and Zhi Zhong. 2013.
\newblock \href {https://aclanthology.org/W13-3516} {Towards robust linguistic
  analysis using {O}nto{N}otes}.
\newblock In \emph{Proceedings of the Seventeenth Conference on Computational
  Natural Language Learning}, pages 143--152, Sofia, Bulgaria. Association for
  Computational Linguistics.

\bibitem[{Raffel et~al.(2020)Raffel, Shazeer, Roberts, Lee, Narang, Matena,
  Zhou, Li, and Liu}]{2020t5}
Colin Raffel, Noam Shazeer, Adam Roberts, Katherine Lee, Sharan Narang, Michael
  Matena, Yanqi Zhou, Wei Li, and Peter~J. Liu. 2020.
\newblock \href {http://jmlr.org/papers/v21/20-074.html} {Exploring the limits
  of transfer learning with a unified text-to-text transformer}.
\newblock \emph{Journal of Machine Learning Research}, 21(140):1--67.

\bibitem[{Rajpurkar et~al.(2016)Rajpurkar, Zhang, Lopyrev, and
  Liang}]{rajpurkar2016squad}
Pranav Rajpurkar, Jian Zhang, Konstantin Lopyrev, and Percy Liang. 2016.
\newblock Squad: 100,000+ questions for machine comprehension of text.
\newblock In \emph{Proceedings of the 2016 Conference on Empirical Methods in
  Natural Language Processing}, pages 2383--2392.

\bibitem[{Saakyan et~al.(2021)Saakyan, Chakrabarty, and
  Muresan}]{saakyan-etal-2021-covid}
Arkadiy Saakyan, Tuhin Chakrabarty, and Smaranda Muresan. 2021.
\newblock \href {https://doi.org/10.18653/v1/2021.acl-long.165} {{COVID}-fact:
  Fact extraction and verification of real-world claims on {COVID}-19
  pandemic}.
\newblock In \emph{Proceedings of the 59th Annual Meeting of the Association
  for Computational Linguistics and the 11th International Joint Conference on
  Natural Language Processing (Volume 1: Long Papers)}, pages 2116--2129,
  Online. Association for Computational Linguistics.

\bibitem[{Shaar et~al.(2020)Shaar, Babulkov, Da~San~Martino, and
  Nakov}]{shaar2020known}
Shaden Shaar, Nikolay Babulkov, Giovanni Da~San~Martino, and Preslav Nakov.
  2020.
\newblock That is a known lie: Detecting previously fact-checked claims.
\newblock In \emph{Proceedings of the 58th Annual Meeting of the Association
  for Computational Linguistics}, pages 3607--3618.

\bibitem[{Shaar et~al.(2021)Shaar, Hasanain, Hamdan, Ali, Haouari, Nikolov,
  Kutlu, Kartal, Alam, Da~San~Martino et~al.}]{shaar2021overview}
Shaden Shaar, Maram Hasanain, Bayan Hamdan, Zien~Sheikh Ali, Fatima Haouari,
  Alex Nikolov, M{\"u}cahid Kutlu, Yavuz~Selim Kartal, Firoj Alam, Giovanni
  Da~San~Martino, et~al. 2021.
\newblock Overview of the clef-2021 checkthat! lab task 1 on check-worthiness
  estimation in tweets and political debates.
\newblock In \emph{CLEF (Working Notes)}.

\bibitem[{Stab and Gurevych(2014)}]{stab2014identifying}
Christian Stab and Iryna Gurevych. 2014.
\newblock Identifying argumentative discourse structures in persuasive essays.
\newblock In \emph{Proceedings of the 2014 Conference on Empirical Methods in
  Natural Language Processing (EMNLP)}, pages 46--56.

\bibitem[{Stab et~al.(2018)Stab, Miller, Schiller, Rai, and
  Gurevych}]{stab2018cross}
Christian Stab, Tristan Miller, Benjamin Schiller, Pranav Rai, and Iryna
  Gurevych. 2018.
\newblock Cross-topic argument mining from heterogeneous sources.
\newblock In \emph{Proceedings of the 2018 Conference on Empirical Methods in
  Natural Language Processing}, pages 3664--3674.

\bibitem[{Strubell et~al.(2019)Strubell, Ganesh, and
  McCallum}]{strubell2019energy}
Emma Strubell, Ananya Ganesh, and Andrew McCallum. 2019.
\newblock Energy and policy considerations for deep learning in {NLP}.
\newblock In \emph{Proceedings of the 57th Annual Meeting of the Association
  for Computational Linguistics}, ACL~'19, pages 3645--3650, Florence, Italy.

\bibitem[{Thorne et~al.(2018)Thorne, Vlachos, Christodoulopoulos, and
  Mittal}]{thorne2018fever}
James Thorne, Andreas Vlachos, Christos Christodoulopoulos, and Arpit Mittal.
  2018.
\newblock Fever: a large-scale dataset for fact extraction and verification.
\newblock In \emph{Proceedings of the 2018 Conference of the North American
  Chapter of the Association for Computational Linguistics: Human Language
  Technologies, Volume 1 (Long Papers)}, pages 809--819.

\bibitem[{Vasileva et~al.(2019)Vasileva, Atanasova, M{\`a}rquez,
  Barr{\'o}n-Cede{\~n}o, and Nakov}]{vasileva2019takes}
Slavena Vasileva, Pepa Atanasova, Llu{\'\i}s M{\`a}rquez, Alberto
  Barr{\'o}n-Cede{\~n}o, and Preslav Nakov. 2019.
\newblock It takes nine to smell a rat: Neural multi-task learning for
  check-worthiness prediction.
\newblock In \emph{Proceedings of the International Conference on Recent
  Advances in Natural Language Processing (RANLP 2019)}, pages 1229--1239.

\bibitem[{Williams et~al.(2018)Williams, Nangia, and
  Bowman}]{williams2018broad}
Adina Williams, Nikita Nangia, and Samuel Bowman. 2018.
\newblock A broad-coverage challenge corpus for sentence understanding through
  inference.
\newblock In \emph{Proceedings of the 2018 Conference of the North American
  Chapter of the Association for Computational Linguistics: Human Language
  Technologies, Volume 1 (Long Papers)}, pages 1112--1122.

\bibitem[{Yin et~al.(2019)Yin, Hay, and Roth}]{yin2019benchmarking}
Wenpeng Yin, Jamaal Hay, and Dan Roth. 2019.
\newblock Benchmarking zero-shot text classification: Datasets, evaluation and
  entailment approach.
\newblock In \emph{Proceedings of the 2019 Conference on Empirical Methods in
  Natural Language Processing and the 9th International Joint Conference on
  Natural Language Processing (EMNLP-IJCNLP)}, pages 3914--3923.

\end{thebibliography}
